\documentclass[twocolumn]{biophys-new}
\usepackage[utf8]{inputenc}
\usepackage[colorlinks,allcolors=cyan!70!black]{hyperref}

\usepackage{xcolor,soul}

\usepackage{blkarray}
\usepackage{xtab,afterpage}
\usepackage{subcaption}
\usepackage[pdftex]{graphicx}
\usepackage{threeparttable}
\usepackage[linesnumbered,ruled]{algorithm2e}
\usepackage{amsmath}
\usepackage{floatrow}
\usepackage[utf8]{inputenc}
\usepackage{xcolor}
\usepackage{url}
\usepackage[nottoc]{tocbibind}
\usepackage{graphicx}
\usepackage[bottom]{footmisc}
\usepackage{amsmath}
\usepackage{enumitem}
\usepackage{multirow}
\usepackage{subcaption} % For subfigures
\usepackage{afterpage} % For placing figure on the next page
\usepackage{caption}  % To customize captions
\usepackage{float}    % For H placement specifier
\usepackage{adjustbox}
\definecolor{mydarkgreen}{RGB}{0,150,0}

\newcommand{\eqcite}[1]{\hyperref[#1]{Eq.~\eqref*{#1}}}

\usepackage{titlesec}
\titlespacing{\section}{0pt}{12pt plus 4pt minus 2pt}{0pt plus 2pt minus 2pt}
\titlespacing{\subsection}{0pt}{12pt plus 4pt minus 2pt}{0pt plus 2pt minus 2pt}
\usepackage{lipsum}

\title{Stability Analysis of Non-Linear Classifiers using Gene Regulatory Neural Network for Biological AI}
\runningtitle{Stability Analysis of Non-Linear Classifier using Gene Regulatory Neural Network for Biological AI} %% For page header

\author[1*]{Adrian Ratwatte}
\author[1,2]{Samitha Somathilaka}
\author[1]{Sasitharan Balasubramaniam}
\author[3, 4]{Assaf A. Gilad}

\runningauthor{Author1 and Author2} %% For page header

\affil[1]{School of Computing, University of Nebraska-Lincoln, 104 Schorr Center, 1100 T Street, Lincoln, NE, 68588-0150, USA.}
\affil[2]{VistaMilk Research Centre, Walton Institute for Information and Communication Systems Science, South East Technological University, Waterford, X91 P20H, Ireland.}
\affil[3]{Department of Chemical Engineering and Materials Science, Michigan State University, East Lansing, Michigan, USA.}
\affil[4]{Department of Radiology, Michigan State University, East Lansing, Michigan, USA.}
\corrauthor[*]{aratwatte2@huskers.unl.edu}

\papertype{Article}
\begin{document}

\begin{frontmatter}
% \textcolor{red}{ $\sim$ 200 words (MAX 300)}
\begin{abstract}
%bacteria, GRN, gene expression, chemical reactions, communication of molecules
The Gene Regulatory Network (GRN) of biological cells governs a number of key functionalities that enables them to adapt and survive through different environmental conditions. Close observation of the GRN shows that the structure and operational principles resembles an Artificial Neural Network (ANN), which can pave the way for the development of Biological Artificial Intelligence. In particular, a gene's transcription and translation process resembles a sigmoidal-like property based on transcription factor inputs. In this paper, we develop a mathematical model of gene-perceptron using a dual-layered transcription-translation chemical reaction model, enabling us to transform a GRN into a Gene Regulatory Neural Network (GRNN). % of a gene revealing its  resemblance to a perceptron, that generate sigmoidal-like molecular concentration dynamics when it reaches a stable equilibrium. 
We perform stability analysis for each gene-perceptron within the fully-connected GRNN sub-network to determine temporal as well as stable concentration outputs that will result in reliable computing performance. We focus on a non-linear classifier application for the GRNN, where we analyzed generic multi-layer GRNNs as well as \emph{E.Coli} GRNN that is derived from trans-omic experimental data. Our analysis found that varying the parameters of the chemical reactions can allow us shift the boundaries of the classification region, laying the platform for  programmable GRNNs that suit diverse application requirements.  

\end{abstract}

% {\textcolor{red}{$\sim$ 111 words (MAX 120)} 
\begin{sigstatement}
{In recent years the significance of artificial intelligence has been steadily rising, driven by the development of numerous algorithms that are applicable across various domains. As we envision a future of "\emph{AI everywhere}", we are faced with the prospects of applying AI into media that is beyond silicon technology, such as wet biological environments. In this study our objective is to propose a paradigm of Biological AI that is built from the gene regulatory process. Realizing a vision of Biological machine AI, can result in novel theranostic applications for disease detection and treatment as well as new bio-hybrid computing systems that integrates biological cells with silicon technology. %this can pave the way for  investigate the perceptron-like behavior exhibited by the transcription and translation gene expressions of bacterial cells at the stable equilibrium, and we evaluate this property through its application in a  non-linear classifier. In addition to the mapping the perceptron onto the dual-layered chemical reaction model of a gene, our research opens up new avenues for future developments in bio-computing and bio-AI machines.   
} 

\end{sigstatement}
\end{frontmatter}

\section*{Introduction}
 %importance of AI, applications, heavy weight light weight, consume lot of energy
In recent years, the field of Artificial intelligence (AI) has developed rapidly resulting in sophisticated learning algorithms that have benefited a plethora of applications (e.g., manufacturing, economics, computer vision, robotics, etc.) \cite{chowdhury2012advantages}.
% , \cite{bishop2006pattern} \cite{strong2016applications}). 
Inspired by the functions of neurons, the ultimate vision of AI is to create human-like intelligence that one day will have a working capacity close to the brain. 
Based on the system applications, AI can be categorized into software or hardware-based. Software-based AI includes various forms of algorithms that depends on their structure as well as training process (e.g., convolutional neural networks {\cite{li2021survey}, recurrent neural networks {\cite{medsker2001recurrent}, where a novel applications is large language models such as Generative Pre-trained Transformer (GPT)   \cite{kasneci2023chatgpt}. %In addition, in the transportation sector, AI agents have been deployed recently to detect and report traffic conditions based on  microscopic traffic data \cite{negenborn2009intelligence}. 
\begin{figure}[ht!] %!t
\centering
\includegraphics[width=0.9\linewidth]{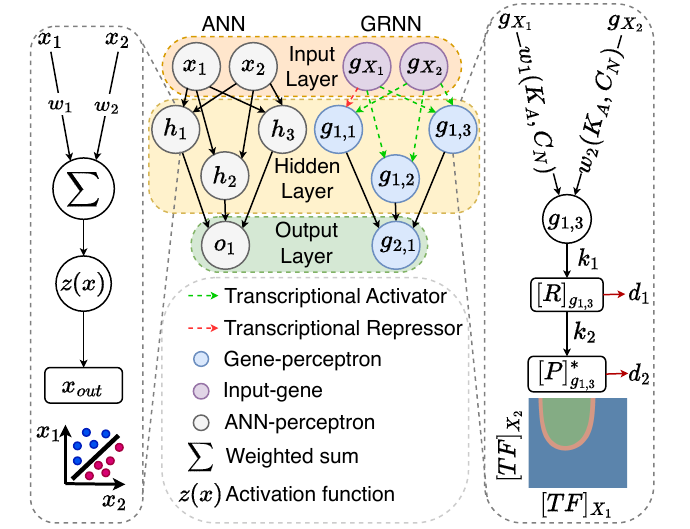}
\caption{Illustration of mapping between components of ANN to GRNN. In this depiction, $w_i$ and $w_i(K_A, C_N)$ represent the weights of a perceptron in ANN and GRNN, respectively, while activation function $z(x)$ is equivalent to a combination of the transcription process of RNA concentration $[R]_i$ as well as translation of maximum-stable protein concentration $[P]_i^*$. The chemical reactions are governed by the transcriptions rate $k_1$, translation rate $k_2$, degradation rate of RNA $d_1$ and degradation rate of protein $d_2$. \vspace{1em}
% In the conventional perceptron, inputs are first weighted prior to taking the weighted summation which is then fed into an activation function. In the gene-perceptron, once the stability at the steady-state is confirmed, maximum-stable protein concentration is calculated for different input-gene concentrations using mathematical models introduced in this study. $K_{i,A}$  is analogous  to the weight and it has a significant impact on the  shifting of the classification area. 
% Notations include weights ($w_1,  w_2$), inputs ($x_1, x_2$), weighted summation ($\sum$), activation function ($z(x)$), output ($x_{out}$) in the ANN-perceptron. $h_1, h_2, h_3$ represent hidden layer nodes and $o_1$ is the output layer node, of the ANN. $g_{X_1}, g_{X_2}$ denote input-genes, $g_{1,i}$ is the 
% $i^{\text{th}}$ hidden layer node, $g_{2,1}$ is the output layer node of the GRNN. $K_{i,A,X_1}, K_{i,A,X_2}$ are transcription factor concentration  corresponding to half maximal RNA concentration of the transcription factors $g_{X_1}, g_{X_2}$, respectively. $[P]_i^*$ signifies the maximum-stable protein concentration for the gene $i$. \textcolor{red}{**NOT finalized yet.}
}
\label{fig:abstract_fig}
\vspace{-0.5em}
\end{figure}

Neuromorphic processors is a hardware-based AI platform that architecturally consists of neurons and synapses constructed from memristor devices that communicate  based on encoded neural spikes. \cite{schuman2022opportunities}.  
%The benefits of both software-based and hardware-based AI include efficiency, multi-functionality, data-based decisions instead of emotional ones and less errors \cite{khanzode2020advantages}.
%However, AI also have several limitations such as lack of explainability \cite{hechler2020limitations}, no guarantee of achieving the optimal solution, issues of potential liability, lack standardized methods to determine the optimal network topology , parameters and activation functions for NN \cite{chowdhury2012advantages}. 
Presently, the vast majority of AI machines are constructed using instruction-encoded circuits and silicon-based semiconductors and nanotechnology \cite{nesbeth2016synthetic}, \cite{akan2016fundamentals}, \cite{akan2023internet}. While this enables more efficient computer systems that have capabilities of learning and computing, it also results in significant challenges such as deployments in wet non-silicon mediums (e.g., biological mediums), as well as utilizing large amounts of energy. %The latter challenge is crucial, when we consider development of algorithms we want to operate close to the efficiency of brain 
\cite{schwenk2005training}.

%sillicon tech
% quiz: can AI be integrated in non sillicon IA, examples bio AI, bio cells as bio AI machines, one line sumary, neural nets from cells
Current research has aimed to address these challenges and one direction taken is through Biological AI, where computing are performed through living biological cells \cite{balasubramaniam2023realizing}, \cite{bi2021survey}. A recent examples is the \emph{DishBrain}, where the system is composed of living neurons that can be trained to play the game of  "\emph{Pong}" on a computer \cite{kagan2022vitro}. 
% Mobile robotics have also been integrated with cultured living neurons to control its operation \cite{warwick2011experiments}.
%Solutions have been proposed that minimizes at least one of these limitations including the challenge of  lacking natural neuronal functionalities.
In other works, ANNs have been programmed into bacterial cells \cite{becerra2022computing}, \cite{li2021synthetic}.
% {\bf[REF - need to add more - i can tell you later}
Similarly, molecular circuits programmed to behave like ANN have also been proposed, and one example is the %For example, Christian et al. showed that phosphorylation/dephosphorylation cycles function as perceptrons, where a stability analysis was conducted for interconnected perceptrons to form a 
Bio-molecular Neural Network (BNN)   \cite{samaniego2021signaling}.
% Similarly, molecular sequestration reactions were implemented as perceptrons and organized into a feed-forward BNN  to solve non-linear classification tasks \cite{moorman2019dynamical}. 
The underlying basis for all these approaches, is the communication of molecules \cite{soldner2020survey}  that operates as part of the chemical reactions to enable computing operations.

%In addition, ANN was used to simulate the behavior of genes including positive, negative feed-back mechanisms where a weight matrix is defined to explain the regulation of one gene on other genes \cite{vohradsky2001neural}. 
%deepsem model GRN based on NN architectiure
%transparent, reduce overfit
%Hantao et al. proposed DeepSEM, a deep generative model that infer gene regulatory network (GRN) using the neural network version of the structural equation model and beta-variational auto-encoder. 

% \sm{I think it is better to define the gap for this study which is the un/stability of GRNNs in this paragraph. So through this study we come up with a model to examine the stability of derived GRNNs.} 
From the perspective of Gene Regulatory Networks (GRN), there has been a connection between its structure and the operation of a ANN. In our recent work \cite{somathilaka2023revealing}, we developed a model that transforms the gene-gene interaction within the GRN using weights, forming a {\bf GRNN} while also exploring the impact of structural changes on the computing capacity. 
% Moreover, weight-loop perceptron and the weight-race perceptron, based on a  chemical perceptron from artificial chemistry were also introduced which were capable of converting weighted summation into output concentrations and weights transform of inputs into outputs \cite{banda2013online}. 
%my work, GRN neural net porperty, AI functionality, subset out of the large, stablity , whole stability, reliable, generic, realGRN, computing classification, 
% significance of results tunin parameters, change area, enginneer cells, depending on the app, say node = gene percetron, weights 
In this study, we investigate the behaviour of a fully-connected GRNN derived from a GRN, focusing on the stability analysis of the gene translation and transcription process during its computing operation. The stability analysis focuses on each perceptron of the GRNN, which we term as   \textbf{gene-perceptron}. Figure \ref{fig:abstract_fig}  illustrates the mapping from ANN to GRNN. 
In a conventional ANN, a perceptron takes multiple inputs ($x_1$ and $x_2$) and computes their weighted summation ($\sum$) that goes through an activation function ($z(x)$). 
% \cite{krogh2008artificial}
%ANNs are commonly employed as a non-linear classifier, with applications extending to various domains. 
% such as the classification of  cardiac arrhythmias \cite{anuradha2008ann}.  
In the context of the GRNN, the weights are represented as Transcription Factors (TF) concentration corresponding to half-maximal RNA concentration ($K_A$) and gene-product copy number ($C_N$), which   individually impact  RNA and protein concentrations. Input-genes ($g_{X_1}$ and $g_{X_2}$) have TFs that binds to the promoter region of the gene-perceptron $g_{1,i}$, which transcribes to RNA ($R_i$) and then translate to protein ($P_i$). This can be considered as a weighted summation, which results in regulatory effects on gene expression within the gene-perceptron. Based on the stability of each gene-perceptron at the steady state, the maximum-stable protein concentration ($[P]^*$), represents the output. %determines the classification area for each gene-perceptron, akin to an activation's function's behavior.  
% Christian et al. demonstrated the  phosphorylation/dephosphorylation cycles  for reliable bio-computing at the stable point \cite{samaniego2021signaling}. Hence, we conduct the stability analysis before determining the classification boundaries of each gene-perceptron.  
   
% Moreover, gene-perceptrons in the hidden layers represents the activation function that also produces TFs for their corresponding successor gene in the network.
% \sm{***I believe, the explanation of gene-perceptron, should be improved something similar to "In contrast, gene-perceptrons in GRNNs take multiple TFs as inputs with varying impacts. This can be considered a weighted summation which lead to regulated expression of the gene-perceptron that resembles a behavior of an activation function." }
We mathematically model chemical reactions of the transcription and translation process of a gene-perceptron, which we term as the dual-layered transcription-translation reaction model (from here on we simply term this as dual-layered chemical reaction model). The dual-layered chemical reaction model can be integrated with trans-omic data model (transcriptome and proteome) and the cellular GRN in order for us to identify active genes for the specific environments, which will be the basis for us to create the GRNN. 
% \sm{The dual-layered chemical reaction model can be integrated with trans-omic data model (transcriptome and proteome) and GRN of the cell species in order for us to obtain active genes for the specific environments.}

% This enables us to derive the weights based on relative expression relationship between the genes, allowing us to form the GRNN \sm{***Do we show how the weights are being derived and used in the model?}. 
Based on this platform, we will perform stability analysis at the steady-state of molecular production (RNA and protein) for the gene-perceptron. Once we prove the stability of the gene-perceptron, as an application we focus on a non-linear classifier relying on the maximum-stable protein concentration for different concentrations of TFs that acts as inputs. To evaluate the model's performance, we analyze two generic multi-layer GRNN networks and an E.Coli GRNN. %based on trans-omic data obtained from wet lab experiments. 
We also show that we can manipulate and shift the classification areas based on different parameter configurations.

%contribution: 1: analysing NN strucutre in GRN discovering
%2: stability analysis 
% 3: app in classification
The contributions of this study can be outlined as follows: 
\begin{itemize}
    \item \textbf{Developing GRNN inspired from ANN structures using dual-layer chemical reaction models:}
    %gene-perceptron, weights, inputs, activation funciton, output 
         Using the dual-layered chemical reaction model, we show that gene transcription and RNA translation process  exhibit a sigmoidal-like molecular concentration dynamics at their stable points. This behavior is governed by the weights, which is a function of gene product copy number and transcription factors TFs concentration corresponding to the half-maximal RNA concentration. %with the inputs represented as transcription factors and the output as the maximum-stable protein concentration.}
          
    \item \textbf{Stability analysis of GRNN:} 
    We developed full mathematical models derived from the chemical reactions and apply Lyapunov's stability analysis for the gene-perceptron to determine stable protein concentration as well as temporal production that will facilitate reliable GRNN computing. %Our analysis is based on concentration as well as temporal stability that will allow an application to determine maximum stability state for the computing operations.  %To the best of our knowledge, we are the pioneers in computing eigenvalues of our system of differential equations to check stability of the gene-perceptron and guaranteed it using Lyapunov's stability theorem in the context of gene expression. Even though, previous work explored stability using aforementioned methods, none has addressed it for gene expression. Mathematically, eigenvalues are used to prove the stability of a system and hence we used it for gene expression.  All derivations are included to find eigenvalues and Lyapunov function including the simulation to illustrate how the Lyapunov stability changes over time enabling us to determine the time that the gene-perceptron achieving stability. 

    \item \textbf{GRNN application for non-linear classifiers:}
    Using the stability analysis, we are able to determine the decision boundaries of the derived GRNNs to classify data within regions of protein concentration output. 
    By varying parameters of the chemical reactions, we demonstrate how the classification area can be shifted,  which can serve as a tool for engineering the GRN for several non-linear classifiers based on the application's requirements.
    
\end{itemize}

\section*{System Modeling} \label{Background}

\begin{figure*}[t!] %!t
\centering
\includegraphics[width=0.8\linewidth]{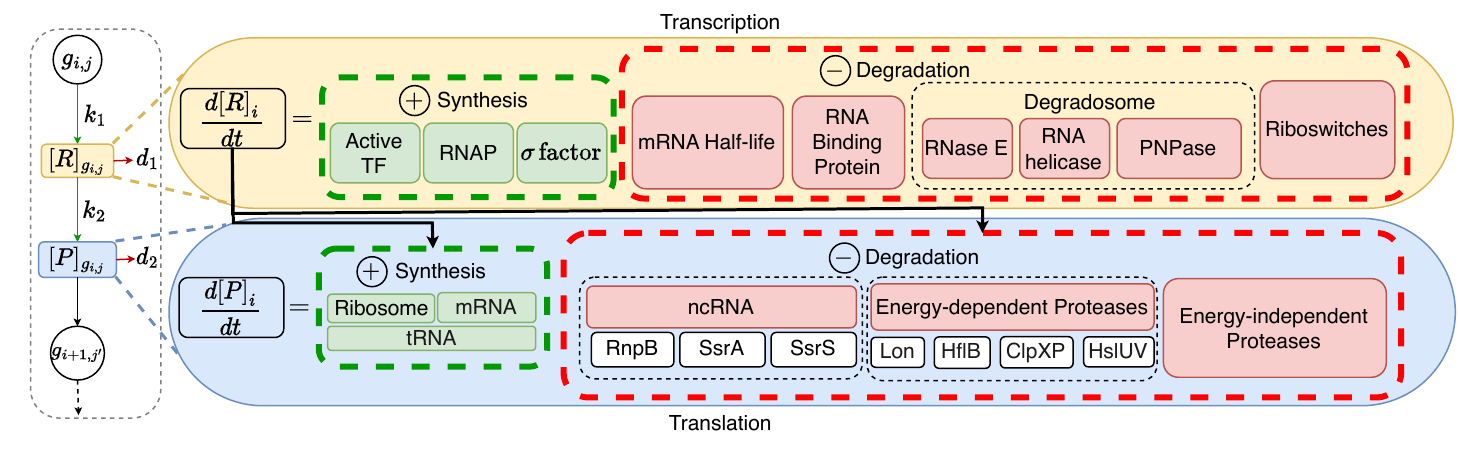}
\caption{Illustration of dual-layered transcription-translation chemical reaction model of the gene-perceptron. Each components corresponds to the synthesis and degradation of RNA and protein for the  $j^{\text{th}}$ gene-perceptron in the $i^{\text{th}}$ layer ($g_{i,j}$) of the GRNN. Here, \emph{RnpB, SsrA} and \emph{SsrS} are examples for non-coding RNA (ncRNA). Examples of energy-dependent proteases include \emph{Lon, HflB, ClpXP} and \emph{HslUV}. Active TF, RNAP, PNPase, RNase E and tRNA corresponds to  active TFs, RNA polymerase,  Polyribonucleotide phosphorylase, Ribonuclease E and transfer RNA, respectively.  }
\label{trans_omic}
\vspace{-0.6em}
\end{figure*}

This section describes the mathematical models for the gene transcription and translation within gene-perceptrons, employing  a dual-layered chemical reaction model (Figure \ref{trans_omic}) that breaks down the steps of the translation and transcription process. 
% This chemical reaction model is composed of molecular processes of production and degradation of transcribed RNAs as well as translated proteins. 
The production of RNAs depends on RNA polymerase, TFs and $\sigma$ factors that binds to the promoter ($Prom$) \cite{wang2021direct}, as well as the dissociation constant ($k_d$). Once the TF binds to the promoters $Prom$, the transcription begins at the rate of $k_1$. This is followed by the RNA degradation at the rate of $d_1$ based on their half-life value \cite{bernstein2002global}, RNA binding proteins \cite{holmqvist2018rna} as well as the degradosome components that includes \emph{RNase E}, \emph{RNA helicase}, as well as \emph{PNPase} \cite{tejada2020bacterial}. Following the transcription of the RNAs is the translation into protein, which occurs at the rate of $k_2$ facilitated by Ribosome and Transfer RNA (tRNA). Once the RNA is translated, the protein molecules start to degrade gradually at the rate of $d_2$. Significant factors that affect the degradation of protein are non-coding RNA, as well as energy-dependent and energy-independent Proteases. Overall, to maintain the concentration stability in the cell, RNAs and protein production are balanced by the degradation process.

By taking the dual-layered chemical reactions model into account, we model the concentration changes at the transcriptome and proteome using mathematical models. These models, enable us to assess the stability of the gene-perceptron expression through the eigenvalue method and determine the  stabilization time using the Lyapunov stability theorem. %Eigen values are determined and hence, we will conclude whether the gene is stable or in-stable based on these Eigen values. If all of these Eigen  values are negative then the gene is considered stable. Conversely,  if any of the Eigen values is not negative, then the gene is not stable. 
After determining if a particular gene-percepton expression is stable, we determine the stability of the entire GRNN. Then, based on the application study, the classification ranges for each gene-perceptron in a network is determined at the equilibrium maximum-stable protein concentration state. Based on the sigmoidal input-output behavior and adjustable threshold, we deduce that gene-perceptrons in the GRNN consist of conventional NN properties.
% Based on the classification areas obtained, we will conclude that  genes in that network act as perceptrons and hence the overall network is functioning as a neural network 
 For the overview of the algorithm mentioned above, please refer to Figure \ref{fig: flow_chart_summary}. 

% The next section introduces the computational models that represent the concentration changes of RNA, protein concentration and the stability analysis of a particular gene-perceptron. The complete derivation of solutions as well as the stability analysis using Lyapunov's theorem can be found in  Appendix A. 
% All notations used in this study, with their corresponding descriptions, are outlined in the Table \ref{notations}. 

% \begin{figure}[ht!] %!t
% \centering
% \includegraphics[width=\linewidth]{figures/basic_model2_horizon.pdf}
% \caption{Flow chart of gene transcription and translation once the transcription factors bind to the promoter region of a gene-perceptron.}
% \label{fig:basic_model_2}
% \end{figure}

\begin{figure}[!t] %!t
\centering
\includegraphics[width=\linewidth]{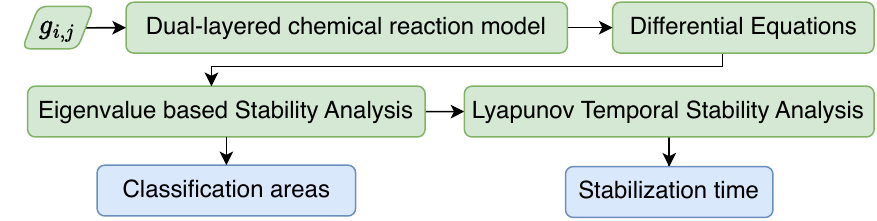}
\caption{Flow chart for the calculation of classification areas as well as stability based on the dual-layred transcription-translation chemical reaction model of each gene-perceptron. \vspace{1em}}
\label{fig: flow_chart_summary}
% \vspace{-0.3em}
\end{figure}

% \begin{table}[t!]
% \caption{Key notations and their descriptions }
% \label{notations}
% \begin{tabular}{ll}
% \hline
% Notation & Description \\ \hline
% $i$ & specific gene-perceptron \\
% $[R]_i$ & RNA concentration (molecules) \\
% $[P]_i$ & Protein concentration (molecules) \\
% $[R]_i^*$ & maximum-stable RNA concentration \\
% $[P]_i^*$ & maximum-stable Protein concentration \\
% $k_{i,1}$ & Transcription rate ($min^{-1}$) \\
% $k_{i,2}$ & Translation rate ($amino \, acids \,sec^{-1}$) \\
% $d_{i,1}$ & Degradation rate of RNA ($min^{-1}$) \\
% $d_{i,2}$ & Degradation rate of Protein ($hour^{-1}$) \\
% $C_{i,N}$ & Gene product copy number (molecules) \\
% $[TF]$ & transcription factor concentration (molecules) \\
% $n$ & Hill coefficient (binding cooperativity) \\
% $K_{i,A}$ & $[TF]$ corresponding to half maximal $[R]_i$ (molecules) \\
% $K_d$ & Dissociation constant \\
% $t$ & time \\ \hline
% \end{tabular}
% \vspace{-0.7em}
% \end{table}

% \vspace{-1.5em}
%network - GRNN, link GRNN, RNA 
\subsection*{Modelling Transcription of a Gene} \label{Modelling_Transcription_Gene}
% description of the process, equation, describe variables, parameter values, how the model is built
In this section, we discuss  transcription and  the corresponding RNA concentration model.
During the transcription process, the RNA polymerase and TFs bind to the promoter region 
% \sm{DNA or region?} 
and then the $\sigma$ factor attaches to the promoter region and unwinds the DNA \cite{cao2020stochastic}. This is followed by the $\sigma$ factor release from the polymerase, allowing for the elongation of the RNA chain.
Based on \cite{santillan2008use}, the concentration change over time $t$ of RNA for a particular gene-perceptron $i$ can be expressed as follows (chemical species are represented using uppercase letters (e.g., $X$), and their corresponding concentration is enclosed within brackets (e.g., $[X]$)) 
 
\begin{equation}
    \dfrac{d[R]_i}{dt}= k_{1_i} C_{N_i} \dfrac{[TF]^n}{K_{A_i}^n + [TF]^n}- d_{1_i} [R]_i. \label{eq:rna_activate}
\end{equation}
The gene-perceptron is activated by the TF, where $[R]_i,$ $k_{1_i}$, $[TF]$, $d_{1_i}$, $n$, $C_{N_i}$ and $K_{A_i}$ are the RNA concentration, transcription rate, concentration of TFs, degradation rate of RNA, Hill coefficient, gene product copy number and TF concentration when the production of RNA is at the half maximal point for gene-perceptron  $i$, respectively.  

Given the initial RNA concentration transcribed by a gene-perceptron is $[R]_i(0)$ (i.e., $[R]_i(t=0)=[R]_i(0)$), the solution of Eq. \ref{eq:rna_activate} is  derived as follows
\begin{equation}
    [R]_i= \dfrac{k_{1_i} C_{N_i}}{d_{1_i}}\left(  \dfrac{[TF]^n}{[TF]^n+K_{A_i}^n} \right) (1-e^{d_{1_i}t}) + [R]_i(0) e^{d_{1_i}t}. \label{eq: sol_rna}
\end{equation}

In contrast, in the event that the gene-perceptron is repressed by the TF, the RNA concentration changes over time $t$  is represented as follows, 
% \sm{the concentration change over time $t$ of RNA will be represented as follows}

\vspace{-0.5em}
\begin{equation}
    \dfrac{d[R]_i}{dt}= k_{1_i} C_{N_i} \dfrac{K_{A_i}^n}{K_{A_i}^n + [TF]^n}- d_{1_i} [R]_i. \label{eq:rna_depress}
\end{equation}

% \vspace{-0.5em}
 Eq. \ref{eq:rna_activate} and  \ref{eq:rna_depress}  is expressed as a mass balance differential equation with the difference between the RNA synthesis,  which is modelled using the Hill function integrated with the degradation process of the RNA \cite{yugi2019rate}, \cite{alon2019introduction}, \cite{thompson2020multiple}. The Hill coefficient $n$ represents the number of  TF molecules that bind simultaneously to the promoter $Prom$ with $K_d$ reaction dissociation constant when the gene-perceptron is transcribing RNA \cite{santillan2008use} and is represented as $Prom + n \,\, TF \stackrel{K_d}{\rightleftharpoons} Prom_{n.TF}$.
The Hill coefficient is critical for the sigmoidal input-output characteristics of the gene-perceptron, as depicted in Figure \ref{fig: Hill_coeff}. According to the plot, we can see that when we increase the hill coefficient, the sigmoidicity increase for the maximum-stable protein concentration ($[P]^*$) over the input-gene concentration ($[TF]$). Thus, when a gene-perceptron possesses a higher hill coefficient, it exhibits more sigmoidal-like behavior. (for our analytical model we consider $n=1$). 
% {\bf**what does cooperativity mean here?}
% According to Figure \ref{fig: Hill_coeff}, increasing {\bf**what is n?} $n$ will increase sigmoidicity of the curve corresponding to the maximum RNA concentration obtained for different transcription factor concentrations. 
%\textcolor{blue}{However, due to lack of wet-lab data to accurately determine cooperativity for genes in the GRN, we assumed that the Hill coefficient value of  $n=1$ for  the simulations conducted on both the  two generic networks and   the subnetwork derivde from the GRN of E.coli.  } {\bf**but in the figure, there is varying n values}. 

% Next section describes the translation of the gene-perceptron and their corresponding mathematical models.  

\begin{figure}[t!] %!t
\centering
\includegraphics[width=0.9\columnwidth]{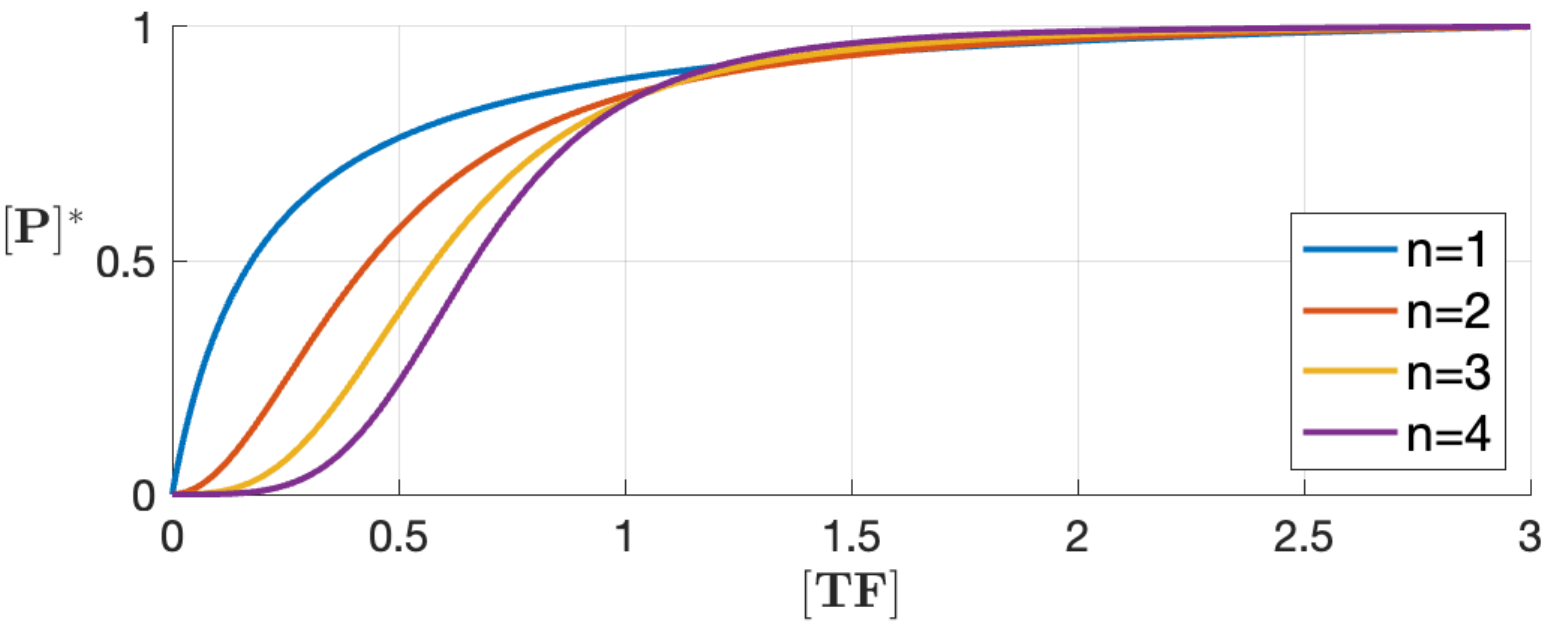}
\caption{Sigmoidicity fluctuations for different Hill coefficients. \vspace{1em}}
\label{fig: Hill_coeff}
\vspace{-0.5em}
\end{figure}

\subsection*{Modelling  Translation of a RNA} \label{RNA_Translation_Models}

In this section, we describe RNA-to-protein translation and associated models.
%the concentration change of protein of a particular gene. 
Initially, the ribosome and tRNAs form a complex that draws the amino acids in the polypeptide chain to attach to the first codon position of the RNA \cite{saito2020translational}. This is followed by the tRNAs adding amino acids one by one to form a polypeptide chain while moving along the RNA \cite{xu2022functions}. Once the stop codon is detected, the polypeptide chain is released, dissociating the ribosome complex from the RNA and forming the protein \cite{marintchev2012fidelity}. This process can be summarized through the protein concentration change over time, and is modelled as follows for a particular gene-perceptron $i$:
\vspace{-0.2em}
\begin{equation}
    \dfrac{d[P]_i}{dt} = k_{2_i}[R]_i - d_{2_i} [P]_i,  \label{eq: prot_change}
\end{equation} 
where $[P]_i, k_{2_i}$ and $d_{2_i}$ are the protein concentration, translation rate and degradation rate of protein for gene-perceptron $i$. Moreover, $[R]_i$ is the concentration of RNA from Eq. \ref{eq:rna_activate}, and the TF activates the gene-perceptron $i$ based on Eq. \ref{eq:rna_depress} if the TF represses the gene-perceptron. Similar to Eq. \ref{eq:rna_activate} and  \ref{eq:rna_depress}, Eq. \ref{eq: prot_change} is modelled based on mass-balance differential equation taking the difference between the RNA produced at the transcriptome level which is translated into protein at the rate of $k_{2_i}$ and the amount of protein that is degraded at the rate of $d_{2_i}$ due to the factors presented in Figure \ref{trans_omic}. 
Provided that the initial protein concentration translated by a RNA for gene-perceptron $i$ is $[P]_i(0)$ (i.e., $[P]_i(t=0)=[P]_i(0)$), the solution of Eq. \ref{eq: prot_change} is given by

\begin{multline}
    [P]_i = \dfrac{k_{1_i}k_{2_i}C_{N_i}}{d_{1_i}} \left( \dfrac{[TF]^n}{[TF]^n+ K_{A_i}^n} \right) \left( \dfrac{1}{d_{2_i}} - \dfrac{e^{d_{1_i}t}}{d_{1_i}+d_{2_i} } \right) \\
    + [R]_i(0) k_{2_i} \left(  \dfrac{e^{d_{1_i}t}}{d_{1_i} + d_{2_i}} \right) 
    +e^{-d_{2_i}t}[P]_i(0)  - e^{-d_{2_i}t} \\ \times [R]_i(0) k_{2_i}  \dfrac{1}{(d_{1_i}+d_{2_i})} 
    - e^{-d_{2_i}t}  \dfrac{k_{1_i}k_{2_i} C_{N_i}}{d_{1_i}} \\ \left(  \dfrac{[TF]^n}{[TF]^n+K_{A_i}^n} \right) 
    \times \left( \dfrac{1}{d_{2_i}} - \dfrac{1}{(d_{1_i}+d_{2_i})} \right).
    \label{eq: sol_prot}
\end{multline}
% At the equilibrium of activator and repressor, 

% \begin{align*}
%     \dfrac{dR}{dt}=0 \xrightarrow[]{} R^* = \dfrac{k_1 C_N}{d_1}. \dfrac{[TF]^n}{K_A^n + [TF]^n} \\
%     \dfrac{dP}{dt}=0 \xrightarrow[]{} P^* = \dfrac{k_2}{d_2}. R^* = \dfrac{k_1 k_2 c_N}{d_1 d_2} \dfrac{[TF]^n}{K_A^n + [TF]^n}\\
%     \xrightarrow[]{} = K_1 \dfrac{[TF]^n}{K_A^n + [TF]^n}\,\,;\,\, \text{where}\,\, K_1=\dfrac{k_1 k_2 c_N}{d_1 d_2}
% \end{align*}
\section*{Methods}
This section introduces the mathematical models for the stability analysis and RNA/Protein concentration changes over time,  and subsequently demonstrates  how to apply these mathematical models in the GRNNs. 
  % \sm{Please give a short intro to the section.}
\subsection*{Gene Expression Stability Analysis} \label{method_stability_analysis}
\vspace{0.3em}

In this section, we discuss the approach towards analyzing the stability of the gene-perceptron expression. % using Eigen values and Lyapunov theorem following similar method used in \cite{samaniego2021signaling} 
%\sm{Please bring this reference into the first stability analysis}. 
Our view of the stability of the gene-perceptron is when the RNA transcription as well as the protein translation concentrations reach maximum over time and remain stable at that level exhibiting a sigmoidal behavior. To confirm the existence of transcription and translation upper bounds, we use eigenvalue-based stability analysis. This, in turn, ensures a stable classification region of the GRNN due to a protein concentration with minimum fluctuations that can result in minimized computing errors. Moreover, another crucial property that should be considered in GRNN computing is the time it takes the GRNN to reach stability, which is investigated using the Lyapunov function in the following sections.

\subsubsection*{Stability of  Gene-Perceptron based on Eigenvalues}

% \textcolor{blue}{Moved the solutions of Eq. \ref{eq:rna_activate} and  \ref{eq: prot_change}  to the System Modelling. }  

The stability of the gene-perceptron is governed by the concentration changes of the gene expression as well as protein translation using the  Jacobian matrix of  Eq. \ref{eq:rna_activate} and  \ref{eq: prot_change}, which enables us to define  the equilibrium point based on the eigenvalues. While we have only considered the case of gene transcription in  Eq. \ref{eq:rna_activate}, our approach is also applicable for repression process defined in Eq. \ref{eq:rna_depress}.
% (please refer to Eq. \ref{eq: define_function_jaco} - \ref{eq: eigen_values} in  Appendix B for the derivations).
Since we are analysing the stability of the gene-perceptron at the equilibrium point, we can represent the  maximum-stable RNA $[R]_i^*$ and protein $[P]_i^*$ concentration as follows:

% The maximum-stable RNA and Protein concentration can be represented as:
\vspace{-1em}
\begin{flalign}
[R]_i^*  &= \dfrac{k_{1_i}C_{N_i}}{d_{1_i}} \left( \dfrac{[TF]^n}{[TF]^n+K_{A_i}^n} \right), \label{eq: max_rna}\\
[P]_i^*  &= \dfrac{k_{1_i}k_{2_i}C_{N_i}}{d_{1_i}d_{2_i}} \left( \dfrac{[TF]^n}{[TF]^n+K_{A_i}^n} \right).\label{eq: max_prot} 
\end{flalign}

The maximum-stable RNA and protein concentrations are determined for different TF concentrations. Additionally, we can vary gene-specific parameters such as $C_{N_i}$ to achieve different non-linear classification ranges \cite{kim2011measuring}, implying that by engineering the cell, we can change its decision-making process.   
 
To determine the eigenvalues of Eq. \ref{eq:rna_activate}  and \ref{eq: prot_change}  at the equilibrium points of Eq. \ref{eq: max_rna} and \ref{eq: max_prot}, we use the  Jacobian matrix given in Eq. \ref{eq: jaco} (please see Appendix).
% $J_i =   
%   \begin{bmatrix}
% -d_{i,1} & 0 \\
% k_{i,2} & -d_{i,2}  \label{eq: jacobian}
% \end{bmatrix}.$
Hence, the eigenvalues are  $\lambda_1 = -d_{1_i}$ and $ \lambda_2=-d_{2_i}$.
Since all the eigenvalues ($\lambda_1$ and $\lambda_2$) are negative, we can conclude that the gene-perceptron  reaches maximum-stable concentration level. 

\subsubsection*{Stability of a Gene-Perceptron using Lyapunov function}
To determine the temporal  stability, we employ Lyapunov stability theorem that is based on the  function ($V([R]_i, [P]_i)$) (from the Appendix Eq. \ref{eq:lyapunov}) which satisfies the necessary conditions: 
$V \left( [R]_i, [P]_i \right)=0$ when $[R]_i= [R]_i^*$ and $[P]_i= [P]_i^*$; where $[R]_i^*$ and $[P]_i^*$ are RNA and protein concentration  at the equilibrium. Additionally, $V \left( [R]_i, [P]_i \right)>0$ due to the quadratic nature of all terms. Finally, we consider the first derivative of Eq. \ref{eq:lyapunov} as given by Eq. \ref{eq:dVdt}, as the last condition to be satisfied for the stability of the gene-perceptron. Then, according to the Lyapunov's theorem, if Eq. \ref{eq:dVdt} is negative, the gene-perceptron is asymptotically stable and if Eq. \ref{eq:dVdt} is less than or equal to zero,  the gene-perceptron is Lyapunov stable (See Eq. \ref{eq:lyapunov} - \ref{eq:dVdt} in the Appendix for the complete derivation). Since it is difficult to directly determine the sign  of the derivative of the Lyapunov function in Eq. \ref{eq:dVdt} (please see the Appendix), we illustrate the temporal fluctuation of Eq. \ref{eq:dVdt} in Figure \ref{lypunov_vs_t}. This provides us the insights into the dynamic stability behavior of the gene-perceptron.   
 % \sm{(I don't think it's a good idea to have results in the Methods section.)} 
 % Between time $0-20 \,$ seconds the derived Lyapunov function is negative exhibiting that the gene-perceptron is asymptotically stable. Subsequently, it approaches zero and remains stable thereafter showing Lyapunov stability \cite{khalil2009lyapunov}. 

% \begin{equation}
%     V([R]_i, [P]_i) = \left( [R]_i - [R]^*_i \right)^2 +  \left( [P]_i - [P]^*_i \right)^2 \label{eq: V}
% \end{equation}
% Equation (\ref{eq: V}) can be further simplified as follows
% \begin{align} 
%     \frac{dV}{dt} &= -\frac{CN^2 \cdot [TF]^{2n} \cdot k_1^2 \cdot e^{(-2t(d_1 + d_2))}}{d_1d_2([TF]^n + K_{i,A}^n)^2(d_1 - d_2)^2}  \nonumber \\ 
%     &\quad \cdot (d_2^3 \cdot e^{(2d_2t)} - 2d_1d_2^2 \cdot e^{(2d_2t)} + d_1^2d_2 \cdot e^{(2d_2t)})  \nonumber \\
%     &\quad \quad + (d_1k_2^2 \cdot e^{(2d_1t)} + d_2k_2^2 \cdot e^{(2d_2t)}) -  \nonumber\\
%     &\quad \quad - (d_1k_2^2 \cdot e^{(t(d_1 + d_2))} + d_2k_2^2 \cdot e^{(t(d_1 + d_2))}). \nonumber
% \end{align}
 % Figure \ref{lypunov_vs_t} represents the stability fluctuation over time based on (\ref{eq:dVdt}). 

% Next section will discuss how the stability and sigmoidal behavior of protein concentration of a gene contributes to the operation of a gene-perceptron. We will discuss how the collection  of gene-perceptrons within a GRNN structure will represent a multi-layer bio-molecular neural network that is used as a non-linear classifier. 

\begin{figure}[ht!] %!t
\centering
\includegraphics[width=\linewidth]{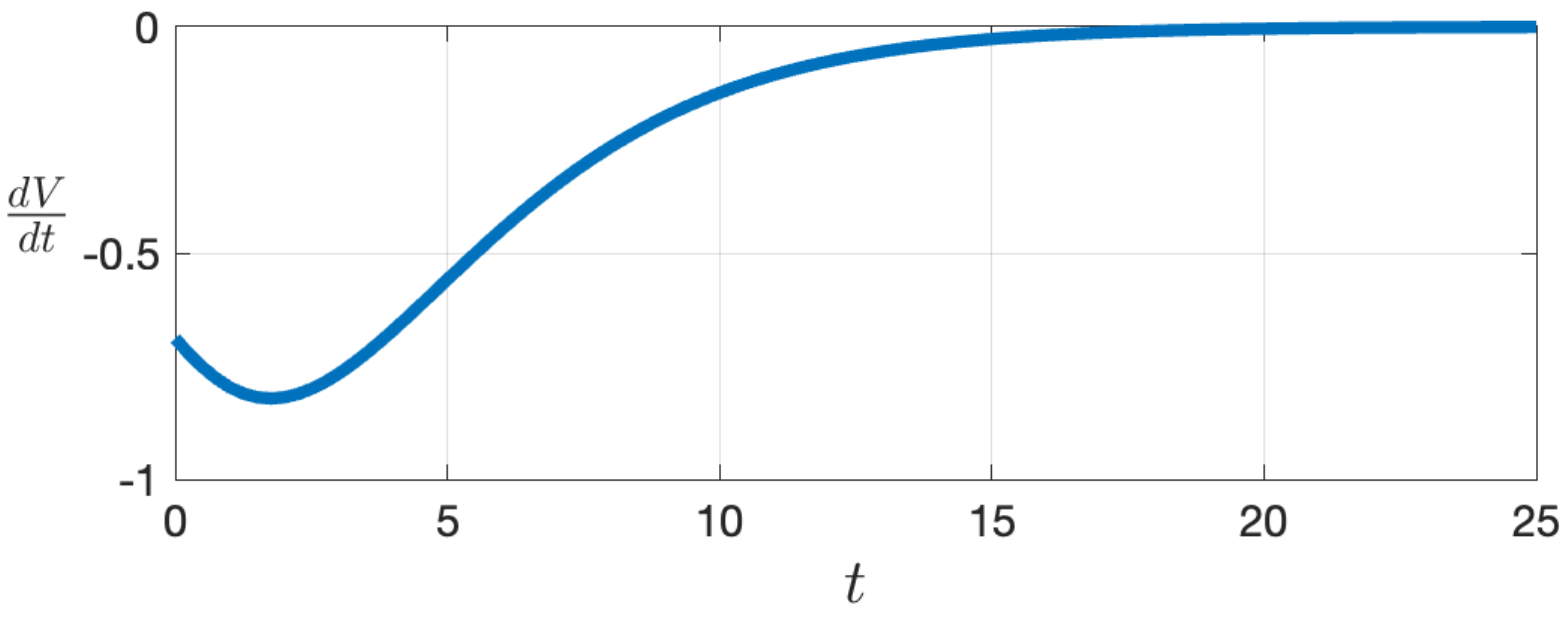}
\caption{Temporal stability of a Gene-perceptron based on  the derivative of the Lyapunov function with respect to time. This shows that the gene-perceptron reaching stability over time.\vspace{-1em}}
\label{lypunov_vs_t}
\vspace{-0.5em}
\end{figure}

% \begin{figure}[ht!] %!t
% \centering
% \includegraphics[width=3.2in]{figures/graph_only.pdf}
% \caption{Output graphs for each node in the sub-network for different input signals }
% \label{TF_vs_P}
% \end{figure}

% \begin{figure}[ht!] %!t
% \centering
% \includegraphics[width=3.4in]{figures/all_in_one.pdf}
% \caption{ Output graphs for each node in the sub-network for different input signals }
% \label{TF_vs_P}
% \end{figure}

% \vspace{-2em}
\subsection*{Gene Regulatory Neural Network Analysis} \label{Network_ana}

While the previous section present the stability analysis of each individual gene-perceptron, they need to be integrated into a GRNN in order to perform the classification operation. We focus on two types of generic multi-layer  GRNNs. In the first network, we consider direct gene relationships within the GRN from the input to the outputs that mimics a multi-layer ANN. In the second case, we consider a Random structured multi-layer GRNN with intermediate gene-perceptrons. 
% \adrian{particularly in cases where no GRNNS with direct relationships are found.} This is the case, where we cannot discover a network with direct gene expression relationships and have to go through an intermediate gene-perceptron. 

%In this section, we will describe computational models for determining the classification areas and conduct stability analysis for two generic  networks and for one network extracted from  gene regulatory network of E.coli.  First, we will consider the generic network without hidden layers. 

\subsubsection*{Multi-layer GRNN} \label{no_hidden}
This GRNN network, which is illustrated in Figure \ref{subnet}, consists of three hidden layer gene-perceptrons ($g_{1,1}, g_{1,2}, g_{1,3}$) and one output layer gene-perceptron ($g_{2,1}$) ($g_{i,j}$ represents the $j^{\text{th}}$ gene-perceptron in $i^{\text{th}}$ layer in the sub-network). The concentrations that is output from layer 1 to layer 2 are $[TF]_{1,1}, [TF]_{1,2}, [TF]_{1,3}$ and $[P]$ is the output from gene-perceptron $g_{2,1}$.
The two input-genes ($g_{X_1} $and $g_{X_2}$) are TFs  with corresponding concentrations ($[TF]_{x_1}$ and $[TF]_{x_2}$), respectively. The RNA  concentration change over time $t$, for the hidden layer gene-perceptrons, based on Eq. \ref{eq:rna_activate}, can be expressed as,
% {\bf**but equation 9 is only RNA and not protein} 
\vspace{-0.4em}
\begin{gather}
    \dfrac{d[R]_i}{dt} = k_{1_i} C_{N_i} \left( \dfrac{[TF]_{x_1}^n}{K_{A_i}^n+ [TF]_{x_1}^n} \right) 
    \cdot \left( \dfrac{[TF]_{x_2}^n}{K_{A_i}^n+ [TF]_{x_2}^n} \right) - d_{1_i} [R]_i, \label{eq:input_activate} 
\end{gather}
for the activators, $i= g_{1,1}, g_{1,2}$.
Since the gene-perceptron $g_{1,3}$ has a repression from gene-perceptron $g_{x_2}$, the changes in the RNA production based on Eq. \ref{eq:rna_depress},  is given by 
\vspace{-0.5em}
\begin{multline}
    \dfrac{d[R]_{g_{1,3}}}{dt} = k_{1_{g_{1,3}}} C_{N_{g_{1,3}}}  \left( \dfrac{[TF]_{x_1}^n}{K_{A_{g_{1,3}}}^n+ [TF]_{x_1}^n} \cdot \dfrac{K_{A_{g_{1,3}}}}{K_{A_{g_{1,3}}}+ [TF]_{x_2}^n}\right)\\ - d_{1_{g_{1,3}}} [R]_{g_{1,3}}.
    \label{eq:input_repress}
\end{multline}
The RNA concentration changes of the output gene-perceptron $g_{2,1}$
% {\bf**where is $g_4$ in the diagram??} 
that consists of TFs from the gene-perceptrons $g_{1,1}, g_{1,2}$ and $g_{1,3}$ with the output protein concentration that contribute as TF concentration ($[TF]_{1,1}= [P]_{g_{1,1}}, [TF]_{1,2}= [P]_{g_{1,2}}$ and $[TF]_{1,3}= [P]_{g_{1,3}}$) to accumulate in order to invoke the expression is given by,
\vspace{-1em}
\begin{multline}
    \dfrac{d[R]_{g_{2,1}}}{dt} = k_{1_{g_{2,1}}} C_{N_{g_{2,1}}}  \left( \dfrac{[TF]_{1,1}^n}{K_{A_{g_{2,1}}}^n+ [TF]_{1,1}^n} \right) \\ \cdot 
      \left( \dfrac{[TF]_{1,2}^n}{K_{A_{g_{2,1}}}^n+ [TF]_{1,2}^n} \right)   \cdot \left( \dfrac{[TF]_{1,3}^n}{K_{A_{g_{2,1}}}^n+ [TF]_{1,3}^n} \right) 
    - d_{ 1_{g_{2,1}} } [R]_{g_{2,1}}.
    \label{eq:g2_1_equation}
\end{multline}

\vspace{-1.5em}
\begin{figure}[ht!] %!t
\centering
\includegraphics[width=0.85\columnwidth]{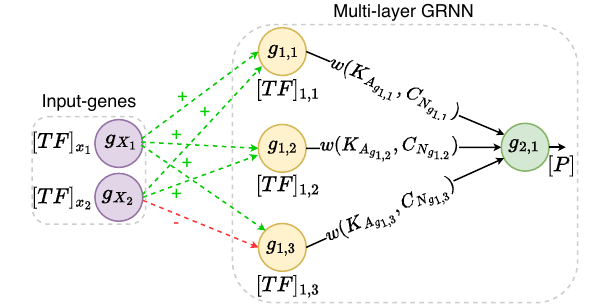}
\caption{Multi-layer GRNN with two-input layer nodes, three hidden-layer gene-perceptrons ($g_{1,1}, g_{1,2}, g_{1,3}$)  and one output layer gene-perceptron ($g_{2,1}$) and their corresponding output concentrations are transcription factors $[TF]_{1,1}, [TF]_{1,2}, [TF]_{1,3} $ and protein concentration $[P]$ respectively. There are two input-genes ($g_{x_1}$, $g_{x_2}$) considered as two TFs  with concentration of $[TF]_{x_1}$ and $[TF]_{x_2}$,  respectively. In this context, $g_{i,j}$ represents the $j^{\text{th}}$ gene-perceptron in $i^{\text{th}}$ layer in the GRNN. Input-gene activators and input-gene repressors are denoted by $(+)$  and  $(-)$ edges, respectively. The weights ($w$) of this GRNN is a function of the TF concentration corresponding to the half-maximal RNA concentration ($K_{A_i}$) and   gene-product copy number ($C_{N_i}$) for the gene-perceptron $i$ represented as $w(K_{A_i}, C_{N_i})$. \vspace{-0.1em}
% \sm {***Please remove weights from the diagrams if you are not using them}. \sm{Can increase the width, so the labels on each array can be shown clearly, and please try to match the font size with the rest. 
% Can include the network name in the figure itself and the caption}
}
\label{subnet}
% \vspace{-0.7em}
\end{figure}

Each of the gene-perceptron also undergoes a translation process. Therefore, the protein concentration change for each gene-perceptron can be modelled using Eq. \ref{eq: prot_change} for $i= g_{1,1}, g_{1,2}, g_{1,3}$ and $g_{2,1}$. The  maximum-stable protein concentration can be derived  by setting Eq. \ref{eq:input_activate} -\ref{eq:g2_1_equation} to zero  to find $[R]_i^*$, which is 
 then plugged into Eq. \ref{eq: prot_change} and set to zero for $i= g_{1,1}, g_{1,2}, g_{1,3}$ and $g_{2,1}$, respectively. 

\vspace{-1em}
\begin{flalign}
   i= g_{1,1}, g_{1,2} \Longrightarrow [P]_i^* = \dfrac{k_{1_i}k_{2_i} C_{N_i}}{d_{1_i}d_{2_i}} \left( \dfrac{[TF]_{x_1}^n}{K_{A_i}^n+ [TF]_{x_1}^n} \right) \nonumber \\
    \times \left( \dfrac{[TF]_{x_2}^n}{K_{A_i}^n+ [TF]_{x_2}^n} \right),  \label{eq: no_hidden_max_prot_g11} 
\end{flalign}
\vspace{-1em}
\begin{flalign}
      [P]_{g_{1,3}}^*  = \dfrac{k_{1_{g_{1,3}}} k_{2_{g_{1,3}}} C_{N_{g_{1,3}}}}{ d_{1_{g_{1,3}}} d_{2_{g_{1,3}}}}  \left( \dfrac{[TF]_{x_1}^n}{K_{A_{g_{1,3}}}^n+ [TF]_{x_1}^n} \right) \nonumber \\ 
    \times \left( \dfrac{K_{A_{g_{1,3}}}}{K_{A_{g_{1,3}}}+ [TF]_{x_2}^n}\right),  \label{eq: no_hidden_max_prot_g13} 
\end{flalign}
\vspace{-1em}
\begin{flalign}
       [P]_{g_{2,1}}^*= \dfrac{k_{1_{g_{2,1}}} k_{2_{g_{2,1}}} C_{N_{g_{2,1}}}}{d_{ 1_{g_{2,1}} } d_{ 2_{g_{2,1}} }}  \left( \dfrac{[TF]_{1,1}^n}{K_{A_{g_{2,1}}}^n+ [TF]_{1,1}^n} \right)  \nonumber \\
   \times  \left( \dfrac{[TF]_{1,2}^n}{K_{A_{g_{2,1}}}^n+ [TF]_{1,2}^n} \right) \left( \dfrac{[TF]_{1,3}^n}{K_{A_{g_{2,1}}}^n+ [TF]_{1,3}^n} \right).
    \label{eq:no_hidden_max_prot_g21}
\end{flalign}

Eq. \ref{eq: no_hidden_max_prot_g11} - \ref{eq:no_hidden_max_prot_g21}, which are the stable concentration quantity of proteins produced, is used to compute the classification areas for each gene-perceptron based on the value of concentration, which is further elaborated in the Results section  as we present a case study.  Subsequently, we apply the approach from the Methods Section   to show the stability of the gene-perceptron in this GRNN. 
The overall stability of the GRNN %based on the derived Lyapunov function (\ref{eq: V}), plug $[R]_i^*$ which is derived from setting (\ref{eq:input_activate})-(\ref{eq:g2_1_equation}) to zero, and    $[P]_i^*$ from Equations (\ref{eq: no_hidden_max_prot_g11})-(\ref{eq:no_hidden_max_prot_g21}).  Finally, find the derivative of this Lyapunov function 
based on the derived Lyapunov function of Eq. \ref{eq:dVdt} (please see Appendix), which can be further expressed for $l$ number of  TFs connected to a gene-perceptron ($i$), is represented as follows
\vspace{-0.5em}
\begin{multline} 
    \frac{dV}{dt} = - \prod_{j=1}^l \frac{C_{N_i}^2 \cdot [TF]_j^{2n} \cdot k_{1_i}^2 \cdot e^{(-2t(d_{1_i} + d_{2_i}))}}{d_{1_i}d_{2_i}([TF]_j^n + K_{A_j}^n)^2(d_{1_i} - d_{2_i})^2} \\
    \times (d_{2_i}^3 \cdot e^{(2d_{2_i}t)} - 2d_{1_i}d_{2_i}^2 \cdot e^{(2d_{2_i}t)} + d_{1_i}^2d_{2_i} \cdot e^{(2d_{2_i}t)}) \\
    + (d_{1_i}k_{2_i}^2 \cdot e^{(2d_{1_i}t)} + d_{2_i}k_{2_i}^2 \cdot e^{(2d_{2_i}t)}) - \\
    - (d_{1_i}k_{2_i}^2 \cdot e^{(t(d_{1_i} + d_{2_i}))} + d_{2_i}k_{2_i}^2 \cdot e^{(t(d_{1_i} + d_{2_i}))}), \label{eq:dVdt_no_hidden} 
\end{multline}
where $[TF]_j$ and $K_{A_j}$ are concentration of $j^{\text{th}}$ TF and   corresponding half maximal RNA concentration for gene-perceptron $i$, respectively. %For the network in Fig. \ref{subnet} , where there are 2 transcription factors for each input layer gene-perceptrons ($g_{1,1}, g_{1,2},g_{1,3}$), we can substitute $l=2$ and other parameters as provided in Table \ref{no_hidden_parameters} into (\ref{eq:dVdt_no_hidden}) to determine the behavior of derivative of the Lyapunov function over time. The following section, demonstrates the models used to compute decision boundaries and conduct stability analysis for the generic network with a hidden layer. 
\begin{table*}[ht!]
\caption{Parameter Configuration for the generic multi-layer GRNN in Fig. \ref{subnet}  }
\label{no_hidden_parameters}
\resizebox{\textwidth}{!}{%
\begin{tabular}{lcccccccccccc}
\hline
Parameter & $C_{N_{g_{1,1}}}$ & $C_{N_{g_{1,2}}}$ & $C_{N_{g_{1,3}}}$ & $C_{N_{g_{2,1}}}$ & $k_{1_{g_{1,1}}}$ & $k_{1_{g_{1,2}}}$ & $k_{1_{g_{1,3}}}$ & $k_{1_{g_{2,1}}}$ & $k_{2_{g_{1,1}}}$ & $k_{2_{g_{1,2}}}$ & $k_{2_{g_{1,3}}}$ & $k_{2_{g_{2,1}}}$ \\
 & $d_{1_{g_{1,1}}}$ & $d_{1_{g_{1,2}}}$ & $d_{1_{g_{1,3}}}$ & $d_{1_{g_{2,1}}}$ & $d_{2_{g_{1,1}}}$ & $d_{2_{g_{1,2}}}$ & $d_{2_{g_{1,3}}}$ & $d_{2_{g_{2,1}}}$ & $K_{A_{g_{1,1}}} (\times10^{-7})$ & $K_{A_{g_{1,2}}}(\times10^{-7})$ & $K_{A_{g_{1,3}}}(\times10^{-7})$ & $K_{A_{g_{2,1}}}(\times10^{-7})$ \\ \hline
Parameter set 1 & 100 & 250 & 500 & 400 & 0.1 & 0.2 & 0.4 & 0.5 & 0.1 & 0.2 & 0.4 & 0.5 \\
 & 0.3 & 0.2 & 0.5 & 0.6 & 0.3 & 0.2 & 0.5 & 0.6 & 500 & 100 & 1000 & 50 \\ \hline
Parameter set 2 & 1 & 2 & 5 & 6 & 0.1 & 0.2 & 0.4 & 0.5 & 0.1 & 0.2 & 0.4 & 0.5 \\
 & 0.3 & 0.2 & 0.5 & 0.6 & 0.3 & 0.2 & 0.5 & 0.6 & $100^*$ & $20^*$ & $10^*$ & $50^*$ \\ \hline \vspace{0em}
\end{tabular}%
}
% Note: The values marked with an asterisk (*) are the parameters that are modified. 
% Units of $C_{i, N}, k_{i,1}, k_{i,2}, d_{i,1}, d_{i,2}$ and $K_{i,A}$ are $molecules, \; sec^{-1}, sec^{-1}, min^{-1}, hour^{-1}$ and $molecules$ respectively for $i= g_{1,1}, g_{1,2}, g_{1,3}$ and $g_{2,1}$.
\vspace{-1.5em}
\end{table*}

\begin{table*}[ht!]
\caption{Parameter configuration for the Random Structured GRNN in Figure \ref{inter_node_1}. \vspace{-0.5em}}
\label{intermediate_node_parameters}
\resizebox{\textwidth}{!}{%
\begin{tabular}{lcccccccccc}
\hline
Parameter & $k_{1_{g_{1,1}}}$ & $k_{1_{g_{1,2}}}$ & $k_{1_{g_{1,3}}}$ & $k_{1_{g_{2,1}}}$ & $k_{1_{g_{3,1}}}$ & $k_{2_{g_{1,1}}}$ & $k_{2_{g_{1,2}}}$ & $k_{2_{g_{1,3}}}$ & $k_{2_{g_{2,1}}}$ & $k_{2_{g_{3,1}}}$ \\
 & $d_{1_{g_{1,1}}}$ & $d_{1_{g_{1,2}}}$ & $d_{1_{g_{1,3}}}$ & $d_{1_{g_{2,1}}}$ & $d_{1_{g_{3,1}}}$ & $d_{2_{g_{1,1}}}$ & $d_{2_{g_{1,2}}}$ & $d_{2_{g_{1,3}}}$ & $d_{2_{g_{2,1}}}$ & $d_{2_{g_{3,1}}}$ \\
 & $C_{N_{g_{1,1}}}$ & $C_{N_{g_{1,2}}}$ & $C_{N_{g_{1,3}}}$ & $C_{N_{g_{2,1}}}$ & $C_{N_{g_{3,1}}}$ & $K_{A_{g_{1,1}}} (\times10^{-7})$ & $K_{A_{g_{1,2}}}(\times10^{-7})$ & $K_{A_{g_{1,3}}}(\times10^{-7})$ & $K_{A_{g_{2,1}}}(\times10^{-7})$ & $K_{A_{g_{3,1}}}(\times10^{-7})$ \\ \hline
Parameter set 1 & 0.1 & 0.2 & 0.4 & 0.8 & 0.5 & 0.1 & 0.2 & 0.4 & 0.7 & 0.5 \\
 & 0.3 & 0.2 & 0.5 & 0.7 & 0.6 & 0.3 & 0.2 & 0.5 & 0.9 & 0.6 \\
 & 1 & 2 & 5 & 10 & 6 & 500 & 100 & 1000 & 50 & 50 \\ \hline
Parameter set 2 & 0.1 & 0.2 & 0.4 & 0.8 & 0.5 & 0.1 & 0.2 & 0.4 & 0.7 & 0.5 \\
 & 0.3 & 0.2 & 0.5 & 0.7 & 0.6 & 0.3 & 0.2 & 0.5 & 0.9 & 0.6 \\
 & 1 & 2 & 5 & 10 & 6 & $50^*$ & $100^*$ & $1000^*$ & $10^*$ & $50^*$ \\ \hline  \vspace{0em}
\end{tabular}%
}   
Note: The values marked with an asterisk (*) are the parameters that are modified. 
Units of $C_{N_i}, k_{1_i}, k_{2_i}, d_{1_i}, d_{2_i}$ and $K_{A_i}$ are $molecules, \; sec^{-1}, sec^{-1}, min^{-1}, hour^{-1}$ and $molecules$ respectively for both tables (Table \ref{no_hidden_parameters} and \ref{intermediate_node_parameters}).
\vspace{-0.5em}
\end{table*}

\subsubsection*{Random Structured GRNN}
% \vspace{-1em}
As described earlier, relationship of gene-perceptrons within a GRN that have common TFs may have intermediate gene-perceptrons within the path of connections. We analyze how this impacts on the overall stability of the GRNN, where the network for this case is presented in Figure \ref{inter_node_1}. In this form of networks, it is necessary to consider the RNA concentration change from the intermediate gene-perceptron ($g_{2,1}$) and its impact on the output layer gene-perceptron ($g_{3,1}$). The expressions for each gene-perceptrons, and their relative TFs from their immediate predecessor, is represented as follows: 
% \vspace{1em}
% \begin{gather}
%     \dfrac{d[R]_{g_{2,1}}}{dt} = k_{{g_{2,1}}, 1} C_{g_{2,1}, N}  \left( \dfrac{[TF]_{1,1}^n}{K_{g_{2,1}, A}^n+ [TF]_{1,1}^n} \right) 
%     - d_{ g_{2,1},1 } [R]_{g_{2,1}}, \label{eq:inter_node}
% \end{gather}

% \vspace{-4em}

\vspace{-1em}
\begin{gather}
  \dfrac{d[R]_{g_{2,1}}}{dt} = k_{1_{g_{2,1}}} C_{N_{g_{2,1}}}  \left( \dfrac{[TF]_{1,1}^n}{K_{A_{g_{2,1}}}^n+ [TF]_{1,1}^n} \right) 
    - d_{ 1_{g_{2,1}}} [R]_{g_{2,1}}, \label{eq:inter_node} \\
    \dfrac{d[R]_{g_{3,1}}}{dt} = k_{1_{g_{3,1}}} C_{N_{g_{3,1}}}  \left( \dfrac{[TF]_{2,1}^n}{K_{A_{g_{3,1}}}^n+ [TF]_{2,1}^n} \right) 
    \cdot  \left( \dfrac{[TF]_{1,2}^n}{K_{A_{g_{3,1}}}^n+ [TF]_{1,2}^n} \right) \nonumber \\ \times \left( \dfrac{[TF]_{1,3}^n}{K_{A_{g_{3,1}}}^n+ [TF]_{1,3}^n} \right) - d_{ 1_{g_{3,1}} } [R]_{g_{3,1}}. \label{eq:inter_node_g31}
\end{gather}

\begin{figure}[ht!] %!t
\centering
\includegraphics[width=\columnwidth]{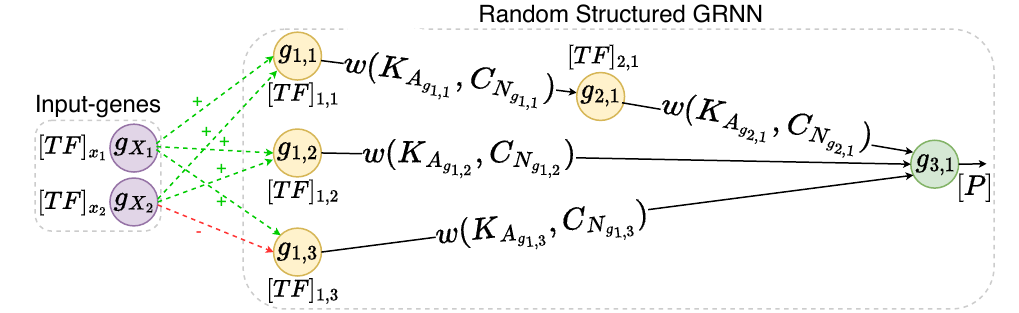}
\caption{Random structured GRNN with 3 input-layer gene-perceptrons ($g_{1,1}, g_{1,2}, g_{1,3}$), one intermediate gene-perceptron ($g_{2,1}$) and one output layer gene-perceptron ($g_{3,1}$). This structure is an extension from the GRNN in Figure 6. \vspace{0.5em}}
\label{inter_node_1}
\vspace{-0.5em}
\end{figure}

Here, the protein concentration from Eq. \ref{eq: sol_prot} can be derived from Eq. \ref{eq:inter_node} (i.e.,  $[TF]_{1,1}=[P]_{1,1}$), since the gene-perceptron $g_{2,1}$ is activated by gene-perceptron $g_{1,1}$. The RNA concentration models behaves similarly to the case without the intermediate gene-perceptron for the gene-perceptrons $g_{1,1}, g_{1,2}$ $g_{1,3}$ and can be derived directly from Eq. \ref{eq:input_activate} and \ref{eq:input_repress}. Using Eq. \ref{eq: prot_change} we can determine the protein concentration change for each gene-perceptron Figure \ref{inter_node_1}. 

Using the  maximum-stable protein concentration derived from Eq. \ref{eq:inter_node} and \ref{eq:inter_node_g31}, we can determine $[R]_i^*$, which is 
 then applied to Eq. \ref{eq: prot_change} and used to determine the maximum-stable value for $i= g_{2,1}$ and $g_{3,1}$. This will result in the following maximum-stable protein production that is represented as follows 

\vspace{-1em}
\begin{flalign}
[P]^*_{g_{2,1}} = \dfrac{k_{1_{g_{2,1}}} k_{2_{g_{2,1}}} C_{N_{g_{2,1}}}}{d_{ 1_{g_{2,1}} } d_{ 2_{g_{2,1}} }}  \left( \dfrac{[TF]_{1,1}^n}{K_{A_{g_{2,1}}}^n+ [TF]_{1,1}^n} \right)  \label{eq: inter_g21},
\end{flalign}
\vspace{-1em}
\begin{flalign}
   [P]^*_{g_{3,1}} = \dfrac{k_{1_{g_{3,1}}} k_{2_{g_{3,1}}} C_{N_{g_{3,1}}} }{d_{ 1_{g_{3,1}} } d_{ 2_{g_{3,1}} }} \left( \dfrac{[TF]_{2,1}^n}{K_{A_{g_{3,1}}}^n+ [TF]_{2,1}^n} \right) \nonumber \\
    \cdot  \left( \dfrac{[TF]_{1,2}^n}{K_{A_{g_{3,1}}}^n+ [TF]_{1,2}^n} \right) \left( \dfrac{[TF]_{1,3}^n}{K_{A_{g_{3,1}}}^n+ [TF]_{1,3}^n} \right) \label{eq: inter_g31}.
\end{flalign}

% \vspace{-0.5em}

We use Eq. \ref{eq: no_hidden_max_prot_g11} to determine $[P]^*_i$ for $i=g_{1,1}$ and $g_{1,2}$, while for  $i=g_{1,3}$ we use Eq. \ref{eq: no_hidden_max_prot_g13}. 
For the stability analysis, Eq. \ref{eq:dVdt_no_hidden} is used with $l=2$ for $g_{1,1}, g_{1,2}$ and $g_{1,3}$, $l=1$ for $g_{2,1}$ and $l=3$ for $g_{3,1}$ corresponding to the number of TFs for each gene-perceptron. %The remaining  parameters are listed in Table \ref{intermediate_node_parameters} . 

% \subsection{Network Extracted from the GRN}

% \begin{multline}
%    [P]^*_{(b1071)} = \dfrac{k_{(b1071), 1} C_{(b1071), N} }{d_{ (b1071),1 } d_{ (b1071),2 }} \\
%    \times \left( \dfrac{[TF]_{(b1891)}^n}{K_{(b1891), A}^n+ [TF]_{(b1891)}^n} \right) \\
%    \times \left( \dfrac{[TF]_{(b1892)}^n}{K_{(b1892), A}^n+ [TF]_{(b1892)}^n} \right)  \label{eq: b1071_real_grn}
% \end{multline}
 
% Commonly, transcription factors bind to the promoter region of the gene and produces a complex which is represented as follows:
% \begin{align*}
%     Prom_{g_1} + n \,\, TF_{x_1} + n\,\, TF_{x_2} \stackrel{K_{d_1}}{\rightleftharpoons} Prom_{n.TF}\\
%     Prom_{g_2} + n \,\, TF_{x_1} + n\,\, TF_{x_2} \stackrel{K_{d_2}}{\rightleftharpoons} Prom_{n.TF}\\
%     Prom_{g_3} + n \,\, TF_{x_1} + n\,\, TF_{x_2} \stackrel{K_{d_3}}{\rightleftharpoons} Prom_{n.TF}\\
%     Prom_{g_4} + n \,\, TF_{1} + n\,\, TF_{2}+ n\,\, TF_{3} \stackrel{K_{d_4}}{\rightleftharpoons} Prom_{n.TF}
% \end{align*}

% \begin{align*}
%     Prom + n \,\, TF_{1} + n\,\, TF_{2} + n\,\, TF_{3} \stackrel{K_{d_4}}{\rightleftharpoons} Prom_{n.TF}
% \end{align*}
% According to  fig. \ref{inter_node_1},  $[TF]_1=[P_{g_1}]$

\section*{Results} \label{results}
In this section, we perform the temporal stability analysis and obtain the classification areas  for the two multi-layer GRNN network topologies (Figures \ref{subnet}, \ref{inter_node_1}) as well as the GRNN derived from \emph{E.Coli} GRN. %presented in \ref{fig:real_net}. 

\subsection*{Multi-layer GRNN} \label{results_no_hidden}

%To obtain the classification area for the single-layer GRNN in Figure \ref{subnet} we will employ mathematical models provided in Section \ref{no_hidden}. Maximum-stable protein concentration ($[P]_i^*$), expressed by (\ref{eq: no_hidden_max_prot_g11}) is for $g_{1,1}$ and  $g_{1,2}$, (\ref{eq: no_hidden_max_prot_g13}) is for the $g_{1,3}$ and (\ref{eq:no_hidden_max_prot_g21}) is for the $g_{2,1}$. 

The temporal stability for each gene-perceptron within the generic multi-layer GRNN is illustrated in Figure \ref{fig:Lypunov_temporal_stability_net1}. This simulation employed the model from Eq. \ref{eq:dVdt_no_hidden}  with the parameter set 1 (Table \ref{no_hidden_parameters}). Gene-perceptrons  $g_{1,1}$ and $g_{1,2}$ initially move away from Lyapunov stability within 5 seconds, indicating a negative trend $ \left( \frac{dV}{dt} < 0 \right)$ in the derivative of the Lyapunov function. 
However, after that it gradually achieved Lyapunov stability maintaining a positive trend and keeping the derivative of the Lyapunov function near zero   $ \left( \frac{dV}{dt} \approx 0 \right)$. In contrast, the gene-perceptron $g_{1,3}$ exhibited positive trend from the beginning reaching Lyapunov stability within 15 seconds due to its distinct repression from the input-gene $g_{X_1}$. The output-layer gene-perceptron ($g_{2,1}$) followed a similar trend as gene-perceptrons $g_{1,1}$ and $g_{1,2}$ attaining Lyapunov stability within the initial 30 seconds because its immediate predecessors are all activators. 

\begin{figure}[ht!] %!t
\centering
\includegraphics[width=\linewidth]{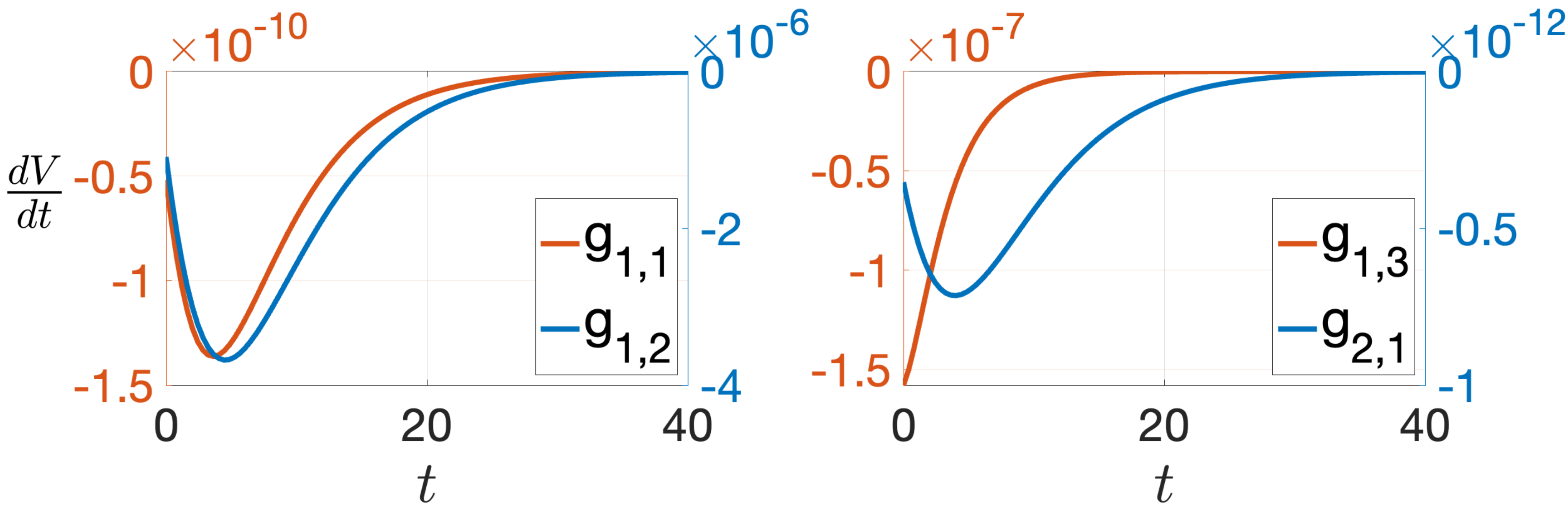}
\vspace{-2em}
\caption{Temporal stability of the gene-perceptrons in the  Multi-layer GRNN.  \vspace{-0.5em}}
\label{fig:Lypunov_temporal_stability_net1}
\end{figure}

\begin{figure*}[ht!]
\centering

\begin{subfigure}{0.8\textwidth}
    \includegraphics[width=\textwidth ]{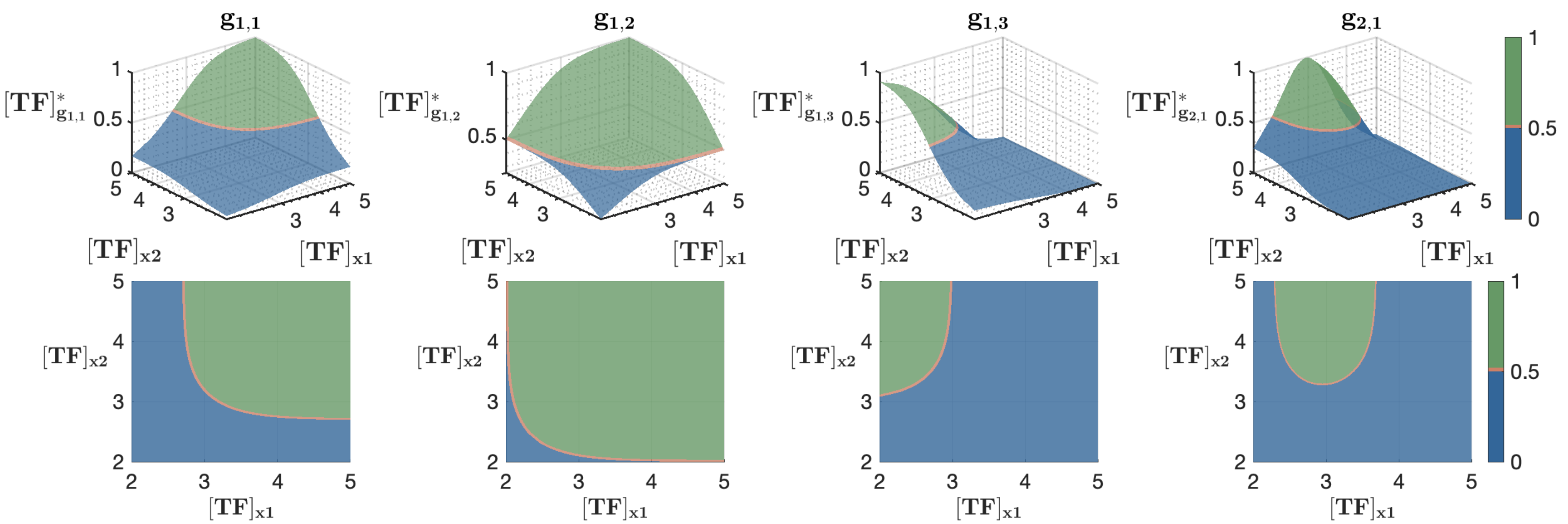}
    \caption{}
    \label{fig:full_net_para_config_1}
\end{subfigure}

\begin{subfigure}{0.8\textwidth}
    \includegraphics[width=\textwidth]{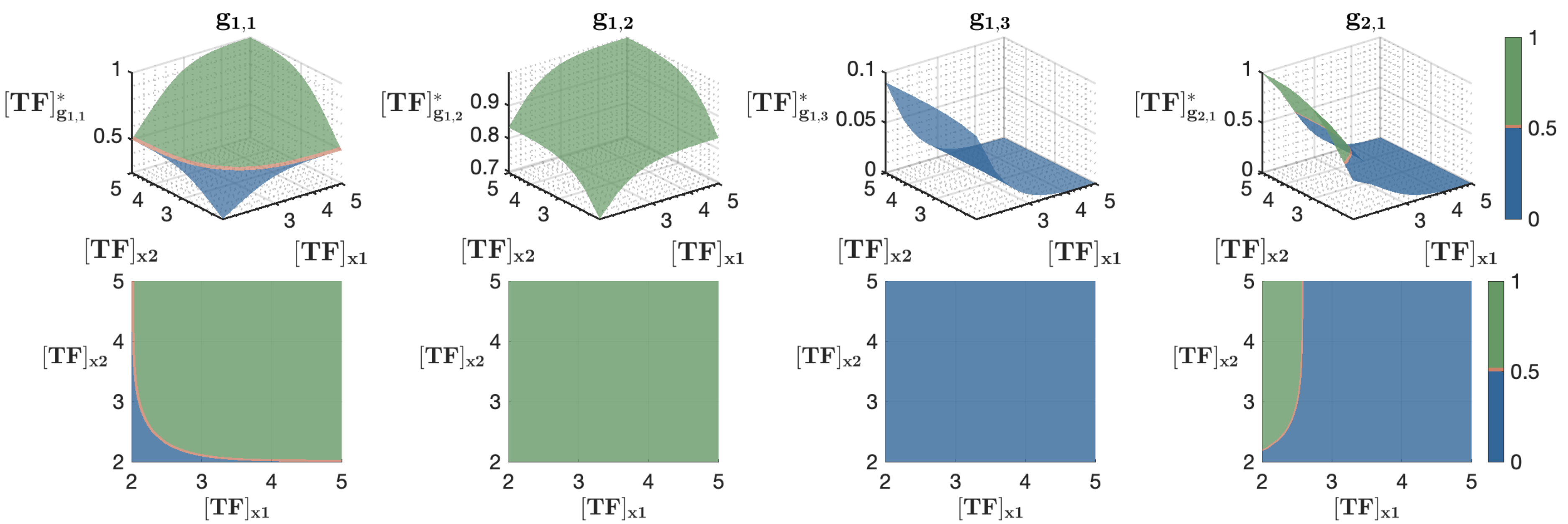}
    \caption{}
    \label{fig:full_net_para_config_2}
\end{subfigure}

% \begin{subfigure}{0.7\textwidth}
%     \includegraphics[width=\textwidth]{figures/para_set-4.pdf}
%     \caption{Parameter set 3}
%     \label{fig:full_net_para_config_3}
% \end{subfigure}
\vspace{-0.5em}
\caption{Parameter configurations for the Multi-layer GRNN depicted in Figure \ref{subnet}. Each graph depicts the classification area of each gene-perceptron and for (a) Parameter set 1, as well as (b) Parameter set 2 ($g_{2,1}$ is the output gene-perceptron that combines all classification areas of gene-perceptrons from the previous layer). \vspace{-0.5em}}
\label{fig:full_net_para_config}

\end{figure*}

Given the gene-perceptron's stability at the equilibrium (Figure \ref{fig:Lypunov_temporal_stability_net1}),  we can use Eq. \ref{eq: no_hidden_max_prot_g11} - \ref{eq:no_hidden_max_prot_g21}  to calculate output protein $[P]_i^*$ for different input concetrations ($[TF]_{x_1}$ and $[TF]_{x_2}$). The calculated output protein $[P]_i^*$ is illustrated over varying input concentrations, highlighting the values above and below the threshold ($[P]^*=0.5$).
% which are the  values that fall between 0.5 and 0.6 
Decision boundaries reflect how the classification areas change based on the edge (activation or repression) connected to the target gene-perceptron and corresponding parameters in  Eq. \ref{eq: no_hidden_max_prot_g11} - \ref{eq:no_hidden_max_prot_g21}. 
The inputs ($[TF]_{x_1}$ and $[TF]_{x_2}$) vary, while parameters like gene product copy number ($C_{N_i}$), transcription rate ($k_{1_i}$), translation rate ($k_{2_i}$), RNA degradation rate ($d_{1_i}$), protein degradation rate ($d_{2_i}$) and TF concentration corresponding to the half maximal RNA concentration ($K_{A_i}$) are kept constant. %are changed for different sets of parameters. 
We consider two  parameters sets to determine the different classification regions, which are presented in Table \ref{no_hidden_parameters}.  
%Simulations obtained for two sets of parameters that yielded significant classifiers are described below. 

% For the parameter set 1, Plasmid copy number ($C_{i,N}$) is 1,2,5 and 6 molecules; transcription rate ($k_{i,1}$) is 0.1, 0.2, 0.4 and 0.5 $min^{-1}$; translation rate ($k_{i,2}$) is 0.1, 0.2, 0.4 and 0.5 $amino \, acids \,sec^{-1}$ ; RNA degradation rate ($d_{i,1}$) is 0.3, 0.2, 0.5 and 0.6 $min^{-1}$; Protein degradation rate ($d_{i,2}$) is 0.3, 0.2, 0.5 and 0.6 $hour^{-1}$;  transcription factor concentration corresponding to half maximal RNA concentration ($K_{i,A}$) is 500, 100, 1000 and 50 (molecules); Hill coefficient, $n=1$ for all parameter sets. 

For the parameters set 1, we obtain the classification areas shown in  Figure \ref{fig:full_net_para_config_1}.
The decision boundary and their top-view for each gene-perceptron are shown in the first  and second row, respectively. 
The gene-perceptron $g_{1,2}$ has the largest classification area above the threshold due its lower  TF concentration corresponding to half maximal RNA concentration $K_{A_i}$,  compared to  
 gene-perceptrons $g_{1,1}$ and $g_{1,3}$. Moreover, the decision boundaries for gene-perceptrons $g_{1,1}$ and $g_{1,2}$ exhibits a similar shape classifying majority of the values above the threshold. In contrast, the gene-perceptron $g_{1,3}$ covers larger area for the values below the threshold since it is repressed by the input-gene $g_{x_2}$. The intersection of classification areas corresponding to hidden layer gene-perceptrons is represented by the output layer gene-perceptron $g_{2,1}$, where the classification area above the threshold is approximately bounded by input concentrations, $2.5 \leq [TF]_{x_1} \leq 3.5$ and $3.4 \leq [TF]_{x_2}$. Due to the significant contribution from gene-perceptrons $g_{1,1}$ and $g_{1,2}$ beyond the threshold, the output layer gene-perceptron $g_{2,1}$ exhibits a rightward shift. %Next, the second parameter configuration will be applied to observe the differences in the results. 

For the parameter set 2 (Table \ref{no_hidden_parameters}), the lower $K_{A_i}$ values have shifted the classification area above the threshold compared to parameter set 1. 
% This shift is evident in  Eq. \ref{eq: no_hidden_max_prot_g11},  $K_{i,A}$ is inversely proportional to the maximum protein concentration ($[P]_i^*$) of the gene-perceptron. 
This shift is evident in Figure \ref{fig:full_net_para_config_2}, particularly for the gene-perceptron $g_{1,2}$, which results in  classifying majority of the values above the threshold. Conversely, for the gene-perceptron $g_{1,3}$, the classification area shifts below the threshold due to the  repression from the input when reducing the  half maximal RNA concentration $K_{A_i}$. The classification range for the gene-perceptron $g_{1,1}$ expands compared to parameter set 1, approximately bounded by $2.3\leq [TF]_{x,1} $ and $2.1\leq [TF]_{x,2} $. Considering all gene-perceptrons, the output layer gene-perceptron $g_{2,1}$ shows a leftward shift in the decision boundary, becoming slightly more linear.  Overall, modifying the  half maximal RNA concentration $K_{A_i}$ can significantly expand the classification area. 
%Next section will describe the simulations obtained for the network with an intermediate gene-perceptron (Fig. \ref{inter_node_1}).
% \vspace{-1em}
\subsection*{Random Structured GRNN} \label{hidden_net_results}

\begin{figure}[t!] %!t
\centering
\includegraphics[width=\linewidth]{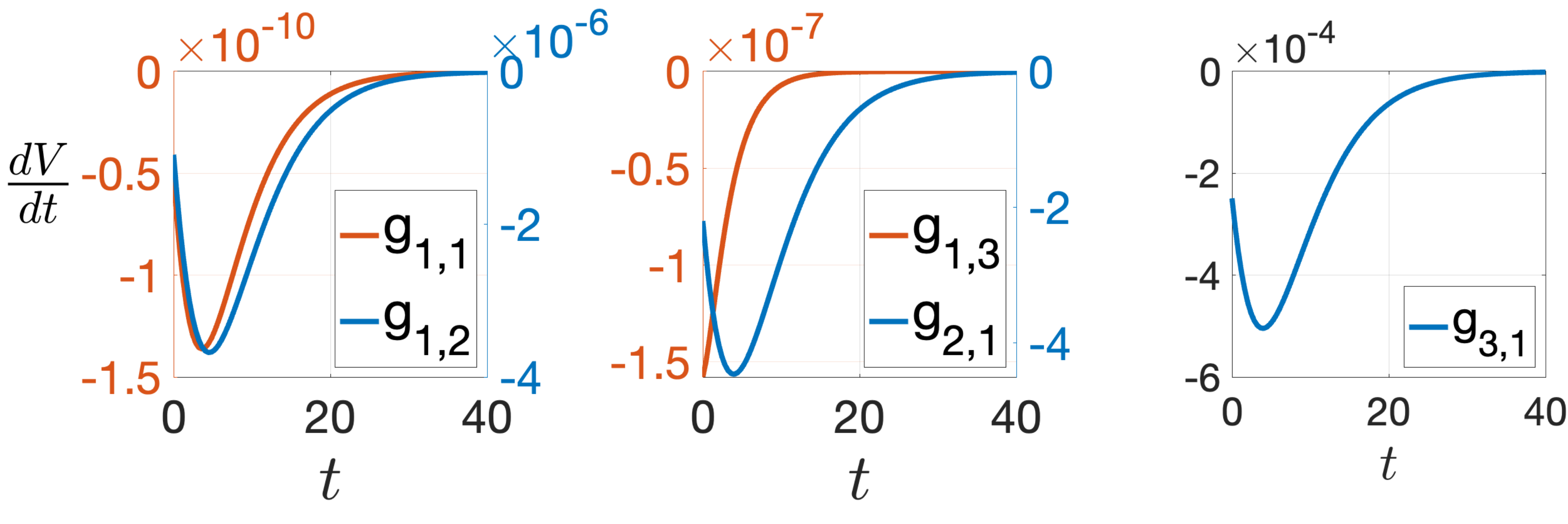}
\caption{Temporal stability of the gene-perceptrons for the Random Structured GRNN. \vspace{-0.5em} }
\label{fig:Lypunov_temporal_stability_net2}
\vspace{-0.5em}
\end{figure}

\begin{figure}[t!] %!t
\centering
\includegraphics[width=0.8\linewidth]{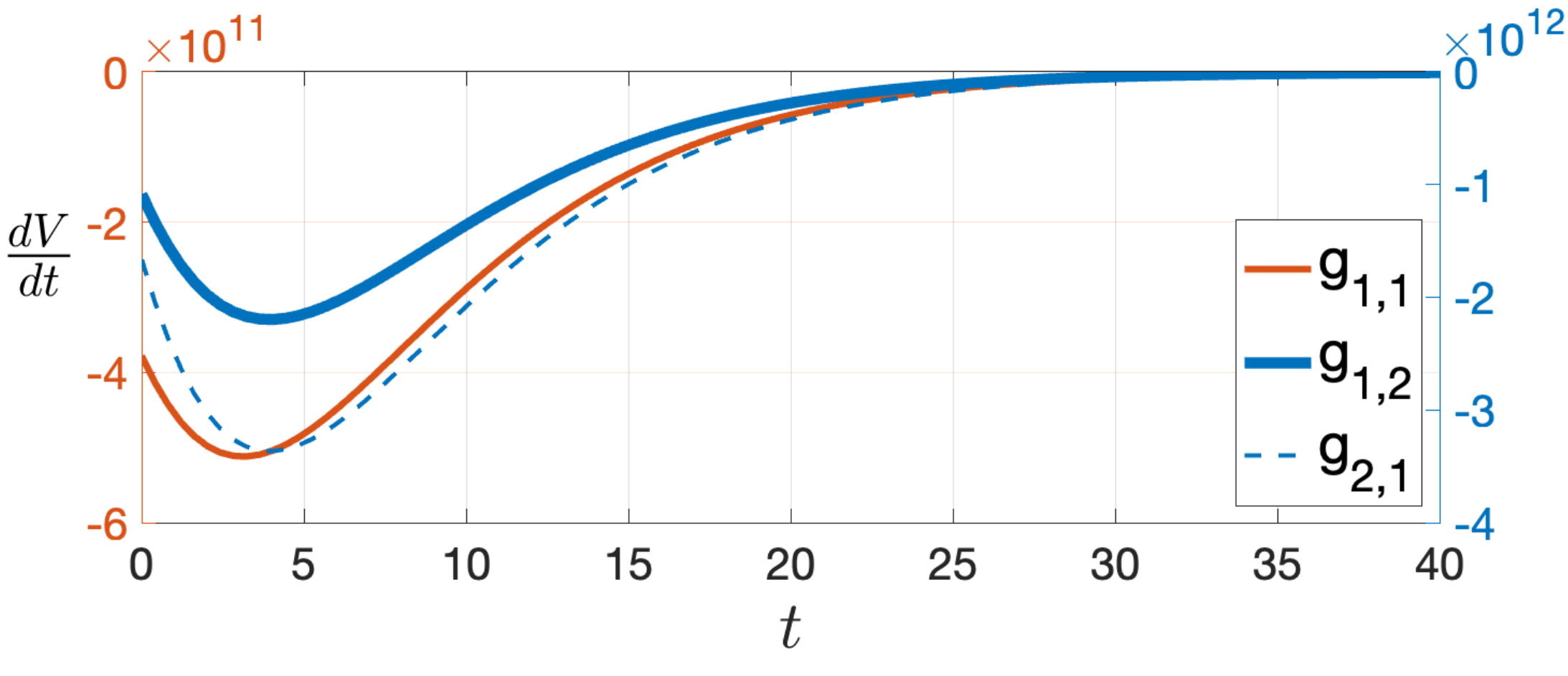}
\caption{Temporal stability for each gene-perceptrons in the \emph{E. coli} GRNN. }
\label{fig:Lypunov_temporal_stability_net3}
\vspace{-0.5em}
\end{figure}

%stable window start measurement 
%intro
This GRNN consists of three hidden layer gene-perceptrons, one intermediate gene-perceptron and one output layer gene-perceptron as illustrated in Figure \ref{inter_node_1}. 
The temporal stability analysis for this GRNN is presented in Figure \ref{fig:Lypunov_temporal_stability_net2} and utilizes Eq. \ref{eq:dVdt_no_hidden}  and the parameter set 1 from Table \ref{intermediate_node_parameters}. Similar to the Figure \ref{fig:Lypunov_temporal_stability_net1},  gene-perceptrons $g_{1,1}, g_{1,2}, g_{3,1}$ and the intermediate gene-perceptron $g_{2,1}$ exhibit consistent stability fluctuations due to  their immediate predecessor being activators. Additionally, gene-perceptron $g_{1,3}$ shows similar stability fluctuation patterns as the gene-perceptron $g_{1,3}$ in the network without the intermediate gene-perceptron and this is because both are being influenced by their repressive predecessors.    

\begin{figure*}[!t] %!t
\centering

\begin{subfigure}{0.9\textwidth}
    \includegraphics[width=\textwidth]{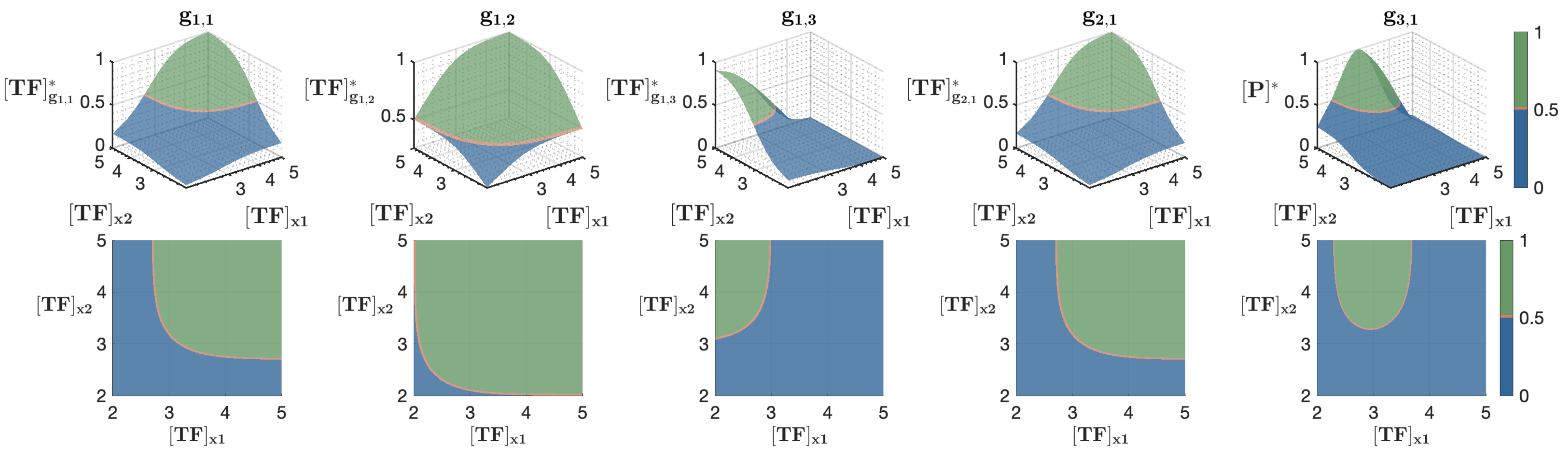}
    \caption{}
    \label{fig:inter_node_para_set_1}
\end{subfigure}

\begin{subfigure}{0.9\textwidth}
    \includegraphics[width=\textwidth]{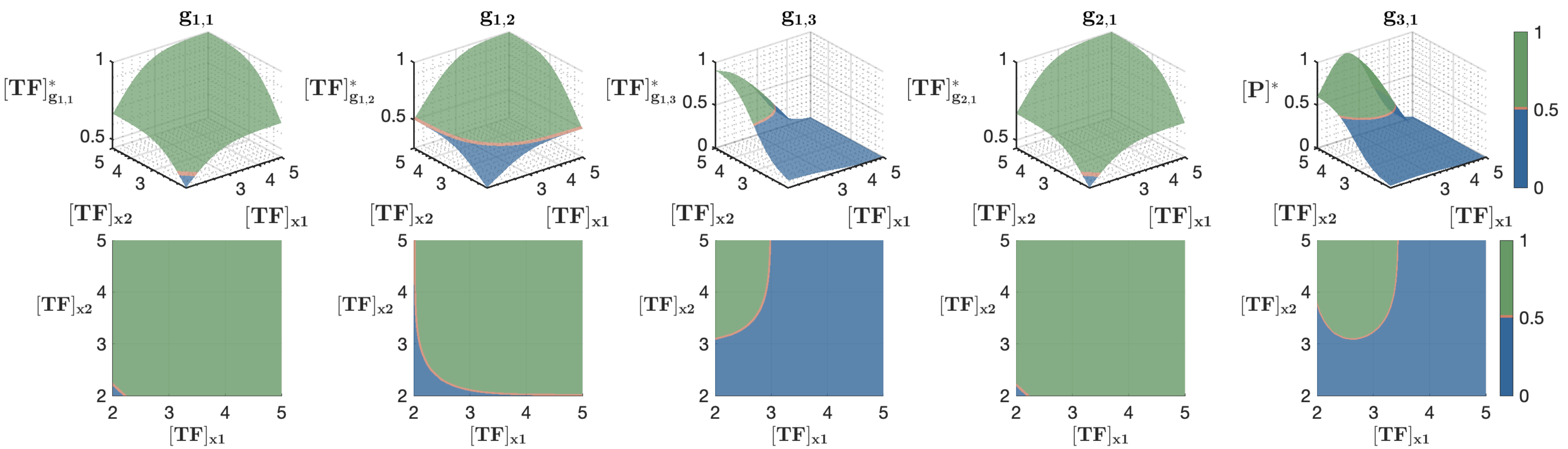}
    \caption{}
    \label{fig:inter_node_para_set_2}
\end{subfigure}
\vspace{-0.5em}
\caption{Parameter configurations for the Random Structured GRNN in Figure \ref{subnet}. Each graph depicts the classification area of each gene-perceptron and for (a) Parameter set 1; (b) Parameter set 2 ($g_{3,1}$ is the output gene-perceptron that combines all classification areas of gene-perceptrons from the previous layer). \vspace{-0.5em}}
\label{fig:inter_node_para_config}

\end{figure*}

Following the temporal stability analysis, we apply Eq. \ref{eq: no_hidden_max_prot_g11} and \ref{eq: no_hidden_max_prot_g13} to  determine the maximum-stable protein concentration ($[P]^*_i$) for the gene-perceptrons $g_{1,1}, g_{1,2}$ and $g_{1,3}$. 
However, unlike the GRNN in Figure \ref{subnet}, Eq. \ref{eq:no_hidden_max_prot_g21} is not used to determine the classification area for the output layer gene-perceptron. 
% Even though the hidden layer is similar in the two GRNNs (Figure \ref{subnet} and \ref{inter_node_1}), the intermediate gene-perceptron and the output layer is different.
Instead, for the computation of  $[P]^*_i$ for the gene-perceptrons $g_{2,1}$ and $g_{3,1}$, both Eq. \ref{eq: inter_g21} and \ref{eq: inter_g31} is employed due to the addition of the intermediate gene-perceptron compared to the multi-layer GRNN in Figure \ref{subnet}. The calculated protein concentration output $[P]^*_i$ values for different input concentrations used to determine the classification area for each gene-perceptron is presented in Figure \ref{fig:inter_node_para_config}. 
We also used two different sets of parameters from Table \ref{intermediate_node_parameters} to analyze different classification areas.
%Transcription rate ($k_{i,1}$), translation rate ($k_{i,2}$), RNA degradation rate ($d_{i,1}$), protein degradation rate ($d_{i,2}$), plasmid copy number ($C_{i,N}$) and transcription factor concentration that gives half maximal RNA concentration ($K_{i,A}$) are assigned while only the significant parameter that contributes to the modification of the decision boundary is changed in parameter set 2. 

%set 1
The parameter set 1 results in the  classification areas shown in Figure \ref{fig:inter_node_para_set_1}. As the gene-perceptron $g_{2,1}$ serves as the intermediate gene-perceptron of $g_{1,1}$, we observe similar classification areas and decision boundaries.
% {\bf**still unclear what its comparable to}.
Additionally, repression from the input-gene $g_{x_1}$ to the gene-perceptron $g_{1,3}$ results in  a distinctive decision boundary,  approximately within the range of $3 \leq [TF]_{x_2}$ and $3 \geq [TF]_{x_1}$. Overall, the gene-perceptron $g_{3,1}$ represents the intersection of the hidden layer gene-perceptrons, with the classification area beyond the threshold bounded by $2.5 \leq [TF]_{x_2} \leq 3.5$ and $3 \geq [TF]_{x_1}$. 
%set 2

In contrast, reducing the TF concentration at the half maximal RNA concentration ($K_{A_i}$) for a gene-perceptron as shown in parameter set 2, alters the classification areas for both gene-perceptron $g_{1,1}$ and its immediate intermediate gene-perceptron $g_{2,1}$, as illustrated in Figure \ref{fig:inter_node_para_set_2}. 
% As presented in Table \ref{intermediate_node_parameters}, we have decreased $K_{i,A}$ for $i=g_{1,1}$ and $ g_{2,1}$ from 500 molecules to 100 molecules and from 50 molecules to 10 molecules, respectively.
The classification area significantly expands  above the threshold, while dropping below it when lowering the TF concentration corresponding to the half-maximal RNA concentration $K_{A_i}$, as it is inversely proportional to the maximum protein concentration $[P]^*_i$ based on  Eqs. \ref{eq:input_activate} and \ref{eq: inter_g21}. Alterations made to gene-perceptron $g_{1,1}$ notably impacts  $g_{2,1}$, the predecessor gene-perceptron in the GRNN.  Other hidden layer gene-perceptrons $g_{1,2}$ and $g_{1,3}$ remain unaffected between parameter sets 1 and 2. Parameter set 2 results in a leftward shift in the classification area of the output layer gene-perceptron $g_{3,1}$ compared to set 1. 
In summary, parameter adjustments leads to shifts in the decision boundary of the output layer gene-perceptrons; with decreased $K_{A_i}$ causing a  leftward shift in the the classification area. %The subsequent section will explain the outcomes of the network extracted from the E.coli gene regulatory network. 

% \footnotesize
% Note: The values marked with an asterisk (*) are the parameters that are modified. Units of $C_{i, N}, k_{i,1}, k_{i,2}, d_{i,1}, d_{i,2}$ and $K_{i,A}$ are $molecules, \; min^{-1}, amino\; acids \; sec^{-1}, min^{-1}, hour^{-1}$ and $molecules$ respectively.

% set 1 - parameters, classification, \\
% set 2 - difference \\
% affects of parameters \\
% parameter table 

% \usepackage{graphicx}
\vspace{-0.5em}
\subsection*{E.Coli GRNN Classification Analysis}  
\begin{figure*}[!ht]
    \centering
    \begin{subfigure}{0.25\textwidth} % Adjust the width to your preference
        \includegraphics[width=\columnwidth]{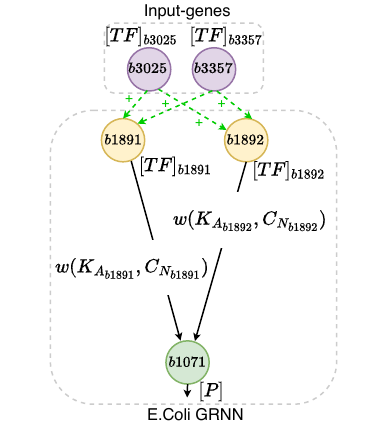}
        \caption{}
        \label{fig:real_net} 
    \end{subfigure}
    \begin{subfigure}{0.6\textwidth} % Adjust the width to your preference
        \includegraphics[width=\columnwidth]{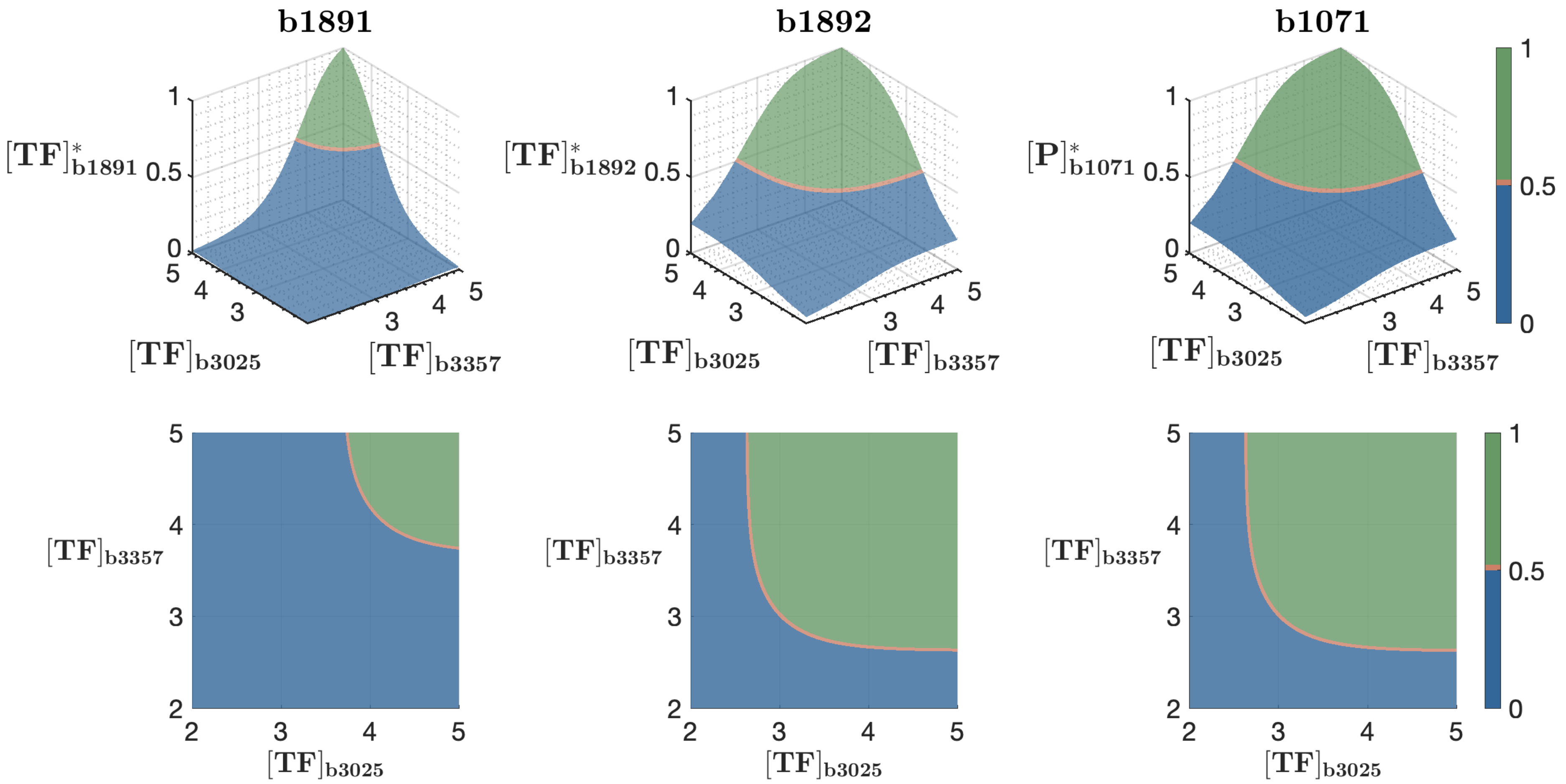}
        \caption{}
        \label{fig:subim2}
    \end{subfigure} 
    \vspace{-0.5em}
    \caption{\emph{E. coli} GRNN classification analysis.  (a) Fully-connected GRNN derived from the E.coli GRN. This network consists of two input-genes ($b3025, b3357$), two hidden layer gene-perceptrons ($b1891$ and $b1892$), and one output layer gene-perceptron ($b1071$). (b) Classification regions of each gene perceptron within the  \emph{E. coli} GRNN, with gene-perceptron $b1071$ as the output. \vspace{-0.5em} }
\end{figure*}

%how you got the subnet
This section demonstrates the classification areas for the  \emph{E.coli} GRNN illustrated in Figure \ref{fig:real_net}, which is extracted from the trans-omic data of \emph{E.coli} GRN \cite{tierrafria2022regulondb}.  
The network consists of two input-genes  ($b3025, b3357$), two hidden layer gene-perceptrons ($b1891$ and $b1892$) and one output layer gene-perceptron ($b1071$) with their corresponding TF concentrations $[TF]_{i}$ for $i=b3025, b3357, b1891$ and $b1892$, and protein concentration $[P]_{b1071}$. In this specific GRNN, all TFs are considered activators.
%We derive  RNA, protein concentration changes and maximum protein concentration for input layer gene-perceptrons using  Eqs. \ref{eq:input_activate}, \ref{eq: prot_change} and \ref{eq: no_hidden_max_prot_g11}. %respectively with $i=b1891$ and $b1892$  and input signals as $x_1=b3025$, $x_2=b3357$. 
For the output layer gene-perceptron ($i=b1071$), we employ   Eqs. \ref{eq:input_activate}, \ref{eq: prot_change} and \ref{eq: no_hidden_max_prot_g11}  with TFs  $x_1=b1891$ and $x_2=b1892$  to calculate RNA, protein concentration change and maximum protein concentration ($[P]^*_i$), respectively using the parameter values in   Table \ref{simulation_val}.

Similar to the previous GRNNs, we based the  stability analysis for this GRNN on Eq. \ref{eq:dVdt_no_hidden}. For the 2 input layer gene-perceptrons ($i=b1891$ and $b1892$), we consider TFs $j= b3025, b3357$, while for the output layer gene-perceptron $i=b1071$, we evaluate stability with the TFs $j= b1891, b1891$ . 
In the previous GRNNs, we found that in Figures \ref{fig:Lypunov_temporal_stability_net1}, \ref{fig:Lypunov_temporal_stability_net2} that the gene-perceptrons with an immediate activator, exhibits a consistent stability fluctuations before reaching Lyapunov stability $ \left( \frac{dV}{dt} \approx 0 \right)$. This is also a similar behaviour with the \emph{E.Coli} GRNN, which is shown in Figure \ref{fig:Lypunov_temporal_stability_net3}, which shows the temporal stability for the gene-perceptrons ($g_{1,1}, g_{1,2}$ and $g_{2,1}$) that is influenced by the immediate activator predecessors displaying uniform  stability. 
Overall, the analysis indicates that  all the gene-perceptrons in the GRNN eventually attained  the Lyapunov stability, ensuring network-wide stability, but with different timing periods.  

\begin{table}[ht!]
\caption{Parameter values used for the \emph{E.coli} GRNN. \vspace{0.5em}}
\label{simulation_val}
\resizebox{\columnwidth}{!}{%
\begin{tabular}{lllll}
\hline
Parameter & \multicolumn{3}{c}{Value} & Ref. \\ \cline{2-4} 
 & b1891 & b1892 & b1071 &  \\ \hline
$k_{1_i}(sec^{-1})$ & $0.05$ & $0.05$ & $0.05$ & \cite{milo2010bionumbers} \\
$k_{2_i}(sec^{-1})$ & 0.05 & 0.05 & 0.05 & \cite{gong2008comparative}, \cite{zhu2019maintenance} \\
$C_{N_i} (molecules)$ & 72 & 122  & 151 & \cite{keseler2017ecocyc} \\
$d_{1_i} (min^{-1})$ & 0.2 & 0.2 & 0.2 & \cite{milo2010bionumbers} \\
$d_{2_i}(hour^{-1})$ & $3.5 \%$ & $3.5 \%$ & $3.5 \%$ & \cite{milo2010bionumbers} \\
$K_{A_i} \left( \times 10^{-7} \right)$ & 75.30 (b3025)  & 71.10 (b3025) & 306 (b1891)  & GSE65244  \\ 
 & 4261.64 (b3357) & 2061.56 (b3357) & 377 (b1892) &  \\
\hline 
\end{tabular}
}
Note:  $K_{A_i}$ of the corresponding TF is given in the table.  
\vspace{-0.8em}
\end{table}

%This extracted network consists of two distinct layers: an input layer containing two gene-perceptrons ($b1891$ and $b1892$) and an output layer with one gene-perceptron ($b1071$). In addition, two transcription factors connected to gene-perceptrons in the input layer, are assumed to be the input signals ($b3357$ and $b3025$). Here, the $b(.)$ represents the gene name of that particular gene-perceptron.  The concentrations of the transcription factors ($[TF]_{b1891}$ and $[TF]_{b1892}$) serve as the input variables for the output layer gene-perceptron which finally outputs the protein concentration. The upcoming section will describe  the use of mathematical models in simulations.  

%math models referring 
Once proving the stability of the GRNN, we ascertain the maximum-stable protein concentration to obtain the classification ranges.  
% {\bf** Its not really characteristics. This word is too high level. you want to talk about reliability of the classification}.  
In order to compute maximum-stable protein concentration ($[P]^*_i$) for gene-perceptrons $i=b1891$ and  $1892 $, we use Eq. \ref{eq: no_hidden_max_prot_g11} with the  replacement of   $x_1$ and $x_2$  by $b3025$ and $b3357$ as input genes. Furthermore, for the computation of output concentrations $[P]^*_i$, concerning gene-perceptron $i=b1071$, Eq. \ref{eq: no_hidden_max_prot_g11} is used with TFs as $x_1=b1891$ and $x_2=b1892$ with the assumption that the Hill coefficient $n$ is equal to 1 in all simulations. Since $K_{A_i}$ is the TF concentration corresponding to the half maximal RNA concentration, there are two $K_{A_i}$ values for each gene-perceptron because each has two TFs, as shown in Figure \ref{fig:real_net}. %To the best of our knowledge, lack of data are available for  $K_{i,A}$ for E.coli. Therefore, 
The time-series data of gene expression levels for \emph{E.coli} was used by first identifying the gene's half maximal expression level $K_{A_i}$ and then finding the  expression level of its TF at that corresponding time point. For the remaining parameters that was obtained from literature as shown in Table \ref{simulation_val}, the average value was used.  
%Finally, using these parameter values and previously mentioned computational models, the classification areas for each gene-perceptron in the GRNN is illustrated and discussed as follows. 

The classification area from our analysis is shown in Figure \ref{fig:subim2}. The classification area of gene-perceptron $b1892$ has expanded towards the left when compared to $b1891$, and this is because the expression level of the half-maximal RNA concentration  $K_{A_i}$ of both TFs ($b3025$ and $b3357$) corresponding to $b1891$ exceed the value of $K_{A_i}$ for $b1892$. 
The classification area above the threshold of $b1892$ is defined within the limits of $[TF]_{b3025} \geq 2.7 $ and $[TF]_{b3357} \geq 2.7 $, in contrast to  $b1891$ which is approximately bounded by $[TF]_{b3025}\geq3.5$ and $[TF]_{b3357}\geq3.8 $.  
Consistent with the decision boundary simulations performed on the two generic multi-layer GRNNs (Figure \ref{fig:full_net_para_config} and \ref{fig:inter_node_para_config}), the output-layer gene-perceptron ($b1071$) of this GRNN also  exhibited a intersection of classification areas driven by the  input-layer gene-perceptrons. In line with this, as gene-perceptron $b1891$ had the majority of its classification area below the threshold and gene-perceptron $b1892$ had the majority above the threshold, the decision boundary of gene-perceptron $b1071$  is approximately bounded by $[TF]_{b3025} \geq 2.9$ and $[TF]_{b3357} \geq 2.9$. Overall, gene-perceptrons within the GRNN derived from  E.coli GRN exhibit  tunable decision boundaries by selecting sub-netowrks from the GRN  at  steady-state and collectively they function as multi-layer GRNN showcasing aspects of biological AI.

\section*{Conclusion} \label{conclusion}
% \vspace{-0.5em}
%summary 
In this study, we introduced a GRNN that can be derived from a cell's GRN and mathematical modelling this for the transcription and translation process, transforming a gene into a gene-perceptron. We also performed stability analysis for the GRNN as it functions as a non-linear classifier. This is based on the eigenvalue method and the Lyapunov's stability theorem, with the latter approach capable of determining the time at which the stability is achieved. 
The classification application was applied to two multi-layer GRNNs as well as a sub-network extracted from the E.coli GRN using trans-omic data. 
%significance 
From the simulation for different parameter settings for the two multi-layer GRNN revealed that the TF concentration at the half maximal gene expression level $K_{A_i}$, has a significant impact on the shifting  of the classification boundary.  Based on the outcomes of the stability analysis and simulations, we can conclude that the GRN exhibits NN properties as the gene-perceptron demonstrated sigmoidal-like behavior for multiple inputs and tunable decision boundary. Further, by engineering living cells it is possible to  obtain desired non-linear classifiers based on our application. 
%limits - parameter values, no subnetworks with hidden layer
Our model has potential to transform GRNs into GRNN when the suitable parameters are established for the dual-layered chemical
reaction model. %Our proposed approach can lead to programmable cells with inherent NN, which can open o %limitations such as lack of data for certain parameters that led us to certain assumptions. 

\section*{Author contributions}
A.R., S.S. and S.B. designed the theoretical framework of the study. The implementation of the analysis was done by A.R. while A.G.  provided the knowledge for the biological aspect of this study. All the authors wrote and reviewed the final manuscript.

\section*{Acknowledgments}
This publication has emanated from research conducted with the financial support of National Science Foundation (NSF) under Grant Number 2316960.

\section*{Declaration of interests}
The authors declare no competing interests.

\section*{Appendix} 
\subsection*{RNA and Protein Concentration Model  } \label{append_A}
To model the RNA and protein concentration change, mass-balance differential equations were used based on Hill function.  
% According to Figure \ref{fig:basic_model_2}, transcription of a gene-perceptron starts with binding of transcription factors and RNA polymerase to the promoter which is modelled using Hill function:
Transcription of a gene-perceptron begins with TF and RNA polymerase binding to the promoter,
% \cite{guo2014transcription},
which is modelled by,
% \vspace{-0.5em}
\begin{gather}
    [Prom.TF] = C_{N_i} \dfrac{[TF]^n}{[TF]^n + K_{A_i}^n}, \label{eq: prmoter_TF_activate}
\end{gather}

where $[TF], n, K_{A_i},  [Prom.TF]$ and $C_{N_i}$ are concentration of TFs, Hill coefficient, TF concentration corresponding to half maximal RNA concentration, complex produced after TFs bind to promoter and gene product copy number, respectively. 
The complex, $Prom.TF$  transcribes into RNA at the rate of $k_{1_i}$ and subsequently RNA degrades at the rate of $d_{1_i}$ which can be modelled as
% \vspace{-1em}
\begin{gather}
    \dfrac{d[R]_i}{dt}= k_{1_i} [Prom.TF]- d_{1_i} [R]_i. \label{eq: rna_activate_2} 
\end{gather}

By plugging Eq. \ref{eq: prmoter_TF_activate} in Eq. \ref{eq: rna_activate_2} we can obtain Eq. \ref{eq:rna_activate}. In contrast, if a gene-perceptron is repressed by a TF,  Eq. \ref{eq: prmoter_TF_activate} can be expressed as
% \vspace{-1.5em}
\begin{equation}
    [Prom.TF]= C_{N_i} \dfrac{K_{A_i}^n}{K_{A_i}^n + [TF]^n}. \label{eq:prmoter_TF_repress}
\end{equation}

 % \vspace{-1em}
Since the initial RNA concentration transcribed by a gene-perceptron is $[R]_i(0)$ (i.e., $[R]_i(t=0)=[R]_i(0)$), the solution of  Eq. \ref{eq:rna_activate} as given by Eq. \ref{eq: sol_rna} can be derived using the integrating factor, $IF=  e^{\int d_{1_i} \,dt}= e^{d_{1_i}t}$ ,
% {\bf** $d_{i,1}$ has not been introduced}
where $t$ and  $d_{1_i}$ are time and RNA degradation rate, respectively.  
Transcribed RNA is then translated into protein at the proteome level. 
To solve the differential equation of protein concentration change for Eq. \ref{eq: prot_change} we can follow 2 steps. {\bf Step 1}: Replacing RNA concentration  ($[R]_i$) in Eq. \ref{eq: prot_change} with the solution obtained for the differential equation of RNA concentration change from Eq. \ref{eq: sol_rna}. {\bf Step 2}: Using the integrating factor ($IF= e^{\int d_{2_i} dt}= e^{d_{2_i}t}$) and initial RNA concentration ($[R]_i(0)$), as well as initial protein concentration $[P]_i(0)$ (i.e., $[P]_i(t=0)=[P]_i(0)$) we can obtain the equation for the protein concentration in Eq. \ref{eq: sol_prot}.  
By setting $\dfrac{d \,[R]_i}{dt}=0$, we can obtain maximum-stable RNA concentration at the steady-state ($[R]_i^*$) expressed by Eq. \ref{eq: max_rna}. In addition, protein concentration at the steady-state ($[P]_i^*$) can be represented by Eq. \ref{eq: max_prot}  which is derived by plugging $\dfrac{d \,[P]_i}{dt}=0$ in Eq. \ref{eq: prot_change}.

% {\bf**say what you are plugging then the equation number
 
% {\bf**what is (3)?} 

% \begin{align}
%     [P]_i = \dfrac{k_{i,1}k_{i,2}C_{i,N}}{d_{i,1}} \left( \dfrac{[TF]^n}{[TF]^n+ K_{i,A}^n} \right) \left( \dfrac{1}{d_{i,2}} - \dfrac{e^{d_{i,1}t}}{d_{i,1}+d_{i,2} } \right) \nonumber \\+ [R]_i(0) k_{i,2} \left(  \dfrac{e^{d{i,1}t}}{d_{i,1} + d_{i,2}} \right)  \nonumber\\+e^{-d_{i,2}t}[P]_i(0) - e^{-d_{i,2}t}[R]_i(0) k_{i,2}  \dfrac{1}{(d_{i,1}+d_{i,2})} \nonumber  \\ - e^{-d_{i,2}t}\dfrac{k_{i,1}k_{i,2} C_{i,N}}{d_{i,1}}\left(  \dfrac{[TF]^n}{[TF]^n+K_{i,A}^n} \right) \left( \dfrac{1}{d_{i,2}} - \dfrac{1}{(d_{i,1}+d_{i,2})} \right) \label{eq: sol_prot}
% \end{align}

% \begin{figure}[ht!] %!t
% \centering
% \includegraphics[width=3 in]{figures/basic_model2.pdf}
% \caption{ Flow chart of gene transcription and translation once the transcription factors bind to the promoter region of a gene  }
% \label{basic_model_2}
% \end{figure}

\subsection*{Determining Gene-perceptron Stability} \label{append_B}
In this section,  we derive the stability of a gene-perceptron
% {\bf**we will not say node, but gene expression - is this correct?} 
using eigenvalues  of differential equations for RNA and protein concentration change (Eq. \ref{eq:rna_activate} and \ref{eq: prot_change}) 
% {\bf**Eigen values of what? what is the eigen value representing - please state, dont just say eigen values}
and using Lypunov's stability theorem. Based on \cite{samaniego2021signaling}, we  applied eigenvalue method to determine the stability in  the gene-perceptrons. 
Suppose $f$ and $g$ are functions of $[R]_i$ and $[P]_i$. Such that, 

\vspace{-1em}
\begin{align}
    \text{Eq.} \ref{eq:rna_activate} \Longrightarrow \dfrac{d \,[R]_i}{dt} &= f \left( [R]_i, [P]_i  \right),\\
    \text{Eq.} \ref{eq: prot_change} \Longrightarrow \dfrac{d \, [P]_i}{dt} &=  g([R]_i,  [P]_i). \label{eq: define_function_jaco}
\end{align}
Then, the  Jacobian matrix for Eqs. \ref{eq:rna_activate} and \ref{eq: prot_change} at the equilibrium point is represented as, 

\vspace{-1em}
\begin{align}
    J_i =   \begin{bmatrix}
\frac{\partial f}{\partial [R]_i} & \frac{\partial f}{\partial [P]_i} \\
\frac{\partial g}{\partial [R]_i} & \frac{\partial g}{\partial [P]_i}
\end{bmatrix}  
=  \begin{bmatrix}
-d_{1_i} & 0 \\
k_{2_i} & -d_{2_i}  \label{eq: jaco} 
\end{bmatrix}, 
\end{align}
for gene-perceptron $i$.
Using the characteristic equation $| J_i- \lambda I | = 0$ we can determine the eigenvalues for the  above Jacobian matrix (Eq. \ref{eq: jaco}) as $\lambda_1 = -d_{1_i}, \lambda_2=-d_{2_i}$.
Hence, all the eigenvalues are negative, indicating that the gene-perceptron is stable,
where $\lambda$ is scalar, $I$ is a $2 \times 2$ identity matrix, $d_{2_i}$ is the protein degradation rate, $d_{1_i}$ is the RNA degradation rate and $k_{2_i}$ is the translation rate. 
We use the Lyapunov function ($V$) to perform the temporal stability analysis defined for the Eqs. \ref{eq:rna_activate} and \ref{eq: prot_change} as follows, 

\vspace{-1em}
\begin{align}
    V \left( [R]_i, [P]_i \right) &= \left( [R]_i - [R]_i^* \right)^2  + \left( [P]_i - [P]_i^* \right)^2 \label{eq:lyapunov}.
\end{align}
According to the Lyapunov's stability theorem, $V \left( [R]_i, [P]_i \right)=0$ when $[R]_i= [R]_i^*$ and $[P]_i= [P]_i^*$, where $[R]_i^*$ and $[P]_i^*$ are RNA and protein concentration  at the equilibrium. It is clear that $V \left( [R]_i, [P]_i \right)>0$, since all terms are quadratic. Finally, we consider the first derivative of Eq. \ref{eq:lyapunov} as the last condition for the stability, which is represented as 

\vspace{-1em}
\begin{align}
    \dot{V}([R]_i,  [P]_i)= \dfrac{dV}{dt} = \frac{\partial V}{\partial [R]_i}. \dfrac{d[R]_i}{dt} + \frac{\partial V}{\partial [P]_i}. \dfrac{d[P]_i}{dt}. \label{eq: derivative_lyapunov}
\end{align} 
By plugging $ \frac{d[R]_i}{dt}$ and $\frac{d[P]_i}{dt}$  from  Eq. \ref{eq:rna_activate} and \ref{eq: prot_change}, differentiating Eq. \ref{eq:lyapunov} with respect to $[R]_i$ and $[P]_i$ to obtain $\frac{\partial V}{\partial [R]_i}$ and $\frac{\partial V}{\partial [P]_i}$ and finally replacing $[R]_i^*, [P]_i^*, [R]_i $ and $[P]_i, $   with Eq. \ref{eq: max_rna}, \ref{eq: max_prot},  \ref{eq: sol_rna} and \ref{eq: sol_prot} we get Eq. \ref{eq: derivative_lyapunov}, which is represented as follows

% \begin{align} 
%    \text{Eq.} \ref{eq: derivative_lyapunov} \Longrightarrow \frac{dV}{dt} &= -\frac{CN^2 \cdot [TF]^{2n} \cdot k_1^2 \cdot e^{(-2t(d_1 + d_2))}}{d_1d_2([TF]^n + K_{i,A}^n)^2(d_1 - d_2)^2}  \nonumber \\ 
%     &\quad \cdot (d_2^3 \cdot e^{(2d_2t)} - 2d_1d_2^2 \cdot e^{(2d_2t)} + d_1^2d_2 \cdot e^{(2d_2t)})  \nonumber \\
%     &\quad \quad + (d_1k_2^2 \cdot e^{(2d_1t)} + d_2k_2^2 \cdot e^{(2d_2t)}) -  \nonumber\\
%     &\quad \quad - (d_1k_2^2 \cdot e^{(t(d_1 + d_2))} + d_2k_2^2 \cdot e^{(t(d_1 + d_2))}), \label{eq:dVdt} 
% \end{align}
\vspace{-1em}
\begin{gather} 
    \text{Eq.} \ref{eq: derivative_lyapunov} \Longrightarrow \frac{dV}{dt} = 
    -\frac{C_{N_i}^2 \cdot [TF]^{2n} \cdot k_{1_i}^2 \cdot e^{(-2t(d_{1_i} + d_{2_i}))}}{d_{1_i}d_{2_i}([TF]^n + K_{A_i}^n)^2(d_{1_i} - d_{2_i})^2} \nonumber \\
    \cdot (d_{2_i}^3 \cdot e^{(2d_{2_i}t)} - 2d_{1_i}d_{2_i}^2 \cdot e^{(2d_{2_i}t)} + d_{1_i}^2d_{2_i} \cdot e^{(2d_{2_i}t)}) \nonumber \\
    + (d_{1_i}k_{2_i}^2 \cdot e^{(2d_{1_i}t)} + d_{2_i}k_{2_i}^2 \cdot e^{(2d_{2_i}t)}) \nonumber \\
    - (d_{1_i}k_{2_i}^2 \cdot e^{(t(d_{1_i} + d_{2_i}))} + d_{2_i}k_{2_i}^2 \cdot e^{(t(d_{1_i} + d_{2_i}))}), \label{eq:dVdt} 
\end{gather}
where we assume initial RNA concentration of zero  ($[R]_i(0)=0$) and  initial protein concentration of zero ($[P]_i(0)=0$). The above equation is used to determine the stability of the gene-perceptron for different parameter configurations. 
% For example, we investigate the stability fluctuation over time as given in Figure \ref{lypunov_vs_t}. 
% Uncomment if using bibtex (default)
\bibliography{sample}

\begin{thebibliography}{35}
\providecommand{\url}[1]{\texttt{#1}}
\providecommand{\urlprefix}{ }

\bibitem[Chowdhury and Sadek(2012)]{chowdhury2012advantages}
Chowdhury, M., and A.~W. Sadek, 2012.
\newblock Advantages and limitations of artificial intelligence.
\newblock \emph{Artificial intelligence applications to critical transportation
  issues} 6:360--375.

\bibitem[Li et~al.(2021{\natexlab{a}})Li, Liu, Yang, Peng, and
  Zhou]{li2021survey}
Li, Z., F.~Liu, W.~Yang, S.~Peng, and J.~Zhou, 2021.
\newblock A survey of convolutional neural networks: analysis, applications,
  and prospects.
\newblock \emph{IEEE transactions on neural networks and learning systems} .

\bibitem[Medsker and Jain(2001)]{medsker2001recurrent}
Medsker, L.~R., and L.~Jain, 2001.
\newblock Recurrent neural networks.
\newblock \emph{Design and Applications} 5:2.

\bibitem[Kasneci et~al.(2023)Kasneci, Se{\ss}ler, K{\"u}chemann, Bannert,
  Dementieva, Fischer, Gasser, Groh, G{\"u}nnemann, H{\"u}llermeier,
  et~al.]{kasneci2023chatgpt}
Kasneci, E., K.~Se{\ss}ler, S.~K{\"u}chemann, M.~Bannert, D.~Dementieva,
  F.~Fischer, U.~Gasser, G.~Groh, S.~G{\"u}nnemann, E.~H{\"u}llermeier, et~al.,
  2023.
\newblock ChatGPT for good? On opportunities and challenges of large language
  models for education.
\newblock \emph{Learning and Individual Differences} 103:102274.

\bibitem[Schuman et~al.(2022)Schuman, Kulkarni, Parsa, Mitchell, Date, and
  Kay]{schuman2022opportunities}
Schuman, C.~D., S.~R. Kulkarni, M.~Parsa, J.~P. Mitchell, P.~Date, and B.~Kay,
  2022.
\newblock Opportunities for neuromorphic computing algorithms and applications.
\newblock \emph{Nature Computational Science} 2:10--19.

\bibitem[Nesbeth et~al.(2016)Nesbeth, Zaikin, Saka, Romano, Giuraniuc, Kanakov,
  and Laptyeva]{nesbeth2016synthetic}
Nesbeth, D.~N., A.~Zaikin, Y.~Saka, M.~C. Romano, C.~V. Giuraniuc, O.~Kanakov,
  and T.~Laptyeva, 2016.
\newblock Synthetic biology routes to bio-artificial intelligence.
\newblock \emph{Essays in biochemistry} 60:381--391.

\bibitem[Akan et~al.(2016)Akan, Ramezani, Khan, Abbasi, and
  Kuscu]{akan2016fundamentals}
Akan, O.~B., H.~Ramezani, T.~Khan, N.~A. Abbasi, and M.~Kuscu, 2016.
\newblock Fundamentals of molecular information and communication science.
\newblock \emph{Proceedings of the IEEE} 105:306--318.

\bibitem[Akan et~al.(2023)Akan, Dinc, Kuscu, Cetinkaya, and
  Bilgin]{akan2023internet}
Akan, O.~B., E.~Dinc, M.~Kuscu, O.~Cetinkaya, and B.~A. Bilgin, 2023.
\newblock Internet of Everything (IoE)-From Molecules to the Universe.
\newblock \emph{IEEE Communications Magazine} .

\bibitem[Schwenk and Gauvain(2005)]{schwenk2005training}
Schwenk, H., and J.-L. Gauvain, 2005.
\newblock Training neural network language models on very large corpora.
\newblock \emph{In} Proceedings of human language technology conference and
  conference on empirical methods in natural language processing. 201--208.

\bibitem[Balasubramaniam et~al.(2023)Balasubramaniam, Somathilaka, Sun,
  Ratwatte, and Pierobon]{balasubramaniam2023realizing}
Balasubramaniam, S., S.~Somathilaka, S.~Sun, A.~Ratwatte, and M.~Pierobon,
  2023.
\newblock Realizing Molecular Machine Learning Through Communications for
  Biological AI.
\newblock \emph{IEEE Nanotechnology Magazine} .

\bibitem[Bi et~al.(2021)Bi, Almpanis, Noel, Deng, and Schober]{bi2021survey}
Bi, D., A.~Almpanis, A.~Noel, Y.~Deng, and R.~Schober, 2021.
\newblock A survey of molecular communication in cell biology: Establishing a
  new hierarchy for interdisciplinary applications.
\newblock \emph{IEEE Communications Surveys \& Tutorials} 23:1494--1545.

\bibitem[Kagan et~al.(2022)Kagan, Kitchen, Tran, Habibollahi, Khajehnejad,
  Parker, Bhat, Rollo, Razi, and Friston]{kagan2022vitro}
Kagan, B.~J., A.~C. Kitchen, N.~T. Tran, F.~Habibollahi, M.~Khajehnejad, B.~J.
  Parker, A.~Bhat, B.~Rollo, A.~Razi, and K.~J. Friston, 2022.
\newblock In vitro neurons learn and exhibit sentience when embodied in a
  simulated game-world.
\newblock \emph{Neuron} 110:3952--3969.

\bibitem[Becerra et~al.(2022)Becerra, Guti{\'e}rrez, and
  Lahoz-Beltra]{becerra2022computing}
Becerra, A.~G., M.~Guti{\'e}rrez, and R.~Lahoz-Beltra, 2022.
\newblock Computing within bacteria: Programming of bacterial behavior by means
  of a plasmid encoding a perceptron neural network.
\newblock \emph{BioSystems} 213:104608.

\bibitem[Li et~al.(2021{\natexlab{b}})Li, Rizik, Kravchik, Khoury, Korin, and
  Daniel]{li2021synthetic}
Li, X., L.~Rizik, V.~Kravchik, M.~Khoury, N.~Korin, and R.~Daniel, 2021.
\newblock Synthetic neural-like computing in microbial consortia for pattern
  recognition.
\newblock \emph{Nature communications} 12:3139.

\bibitem[Samaniego et~al.(2021)Samaniego, Moorman, Giordano, and
  Franco]{samaniego2021signaling}
Samaniego, C.~C., A.~Moorman, G.~Giordano, and E.~Franco, 2021.
\newblock Signaling-based neural networks for cellular computation.
\newblock \emph{In} 2021 American Control Conference (ACC). IEEE, 1883--1890.

\bibitem[S{\"o}ldner et~al.(2020)S{\"o}ldner, Socher, Jamali, Wicke,
  Ahmadzadeh, Breitinger, Burkovski, Castiglione, Schober, and
  Sticht]{soldner2020survey}
S{\"o}ldner, C.~A., E.~Socher, V.~Jamali, W.~Wicke, A.~Ahmadzadeh, H.-G.
  Breitinger, A.~Burkovski, K.~Castiglione, R.~Schober, and H.~Sticht, 2020.
\newblock A survey of biological building blocks for synthetic molecular
  communication systems.
\newblock \emph{IEEE Communications Surveys \& Tutorials} 22:2765--2800.

\bibitem[Somathilaka et~al.(2023)Somathilaka, Balasubramaniam, Martins, and
  Li]{somathilaka2023revealing}
Somathilaka, S.~S., S.~Balasubramaniam, D.~P. Martins, and X.~Li, 2023.
\newblock Revealing Gene Regulation-Based Neural Network Computing in Bacteria.
\newblock \emph{Biophysical Reports} .

\bibitem[Wang et~al.(2021)Wang, Dong, Sokac, Golding, and Xu]{wang2021direct}
Wang, J., Y.~Dong, A.~M. Sokac, I.~Golding, and H.~Xu, 2021.
\newblock Direct Quantification of Gene Regulation by Transcription-Factor
  Binding at an Endogenous Gene Locus.
\newblock \emph{Biophysical Journal} 120:260a.

\bibitem[Bernstein et~al.(2002)Bernstein, Khodursky, Lin, Lin-Chao, and
  Cohen]{bernstein2002global}
Bernstein, J.~A., A.~B. Khodursky, P.-H. Lin, S.~Lin-Chao, and S.~N. Cohen,
  2002.
\newblock Global analysis of mRNA decay and abundance in Escherichia coli at
  single-gene resolution using two-color fluorescent DNA microarrays.
\newblock \emph{Proceedings of the National Academy of Sciences} 99:9697--9702.

\bibitem[Holmqvist and Vogel(2018)]{holmqvist2018rna}
Holmqvist, E., and J.~Vogel, 2018.
\newblock RNA-binding proteins in bacteria.
\newblock \emph{Nature Reviews Microbiology} 16:601--615.

\bibitem[Tejada-Arranz et~al.(2020)Tejada-Arranz, de~Crecy-Lagard, and
  de~Reuse]{tejada2020bacterial}
Tejada-Arranz, A., V.~de~Crecy-Lagard, and H.~de~Reuse, 2020.
\newblock Bacterial RNA degradosomes: molecular machines under tight control.
\newblock \emph{Trends in biochemical sciences} 45:42--57.

\bibitem[Cao et~al.(2020)Cao, Filatova, Oyarz{\'u}n, and
  Grima]{cao2020stochastic}
Cao, Z., T.~Filatova, D.~A. Oyarz{\'u}n, and R.~Grima, 2020.
\newblock A stochastic model of gene expression with polymerase recruitment and
  pause release.
\newblock \emph{Biophysical Journal} 119:1002--1014.

\bibitem[Santill{\'a}n(2008)]{santillan2008use}
Santill{\'a}n, M., 2008.
\newblock On the use of the Hill functions in mathematical models of gene
  regulatory networks.
\newblock \emph{Mathematical Modelling of Natural Phenomena} 3:85--97.

\bibitem[Yugi et~al.(2019)Yugi, Ohno, Krycer, James, and Kuroda]{yugi2019rate}
Yugi, K., S.~Ohno, J.~R. Krycer, D.~E. James, and S.~Kuroda, 2019.
\newblock Rate-oriented trans-omics: integration of multiple omic data on the
  basis of reaction kinetics.
\newblock \emph{Current Opinion in Systems Biology} 15:109--120.

\bibitem[Alon(2019)]{alon2019introduction}
Alon, U., 2019.
\newblock An introduction to systems biology: design principles of biological
  circuits.
\newblock CRC press.

\bibitem[Thompson et~al.(2020)Thompson, Marinelli, Bertram, Sherman, and
  Satin]{thompson2020multiple}
Thompson, B.~M., I.~Marinelli, R.~Bertram, A.~Sherman, and L.~S. Satin, 2020.
\newblock Multiple Feedback Mechanisms Underlying Beta Cell Secretory
  Oscillations.
\newblock \emph{Biophysical Journal} 118:562a.

\bibitem[Saito et~al.(2020)Saito, Green, and Buskirk]{saito2020translational}
Saito, K., R.~Green, and A.~R. Buskirk, 2020.
\newblock Translational initiation in E. coli occurs at the correct sites
  genome-wide in the absence of mRNA-rRNA base-pairing.
\newblock \emph{Elife} 9:e55002.

\bibitem[Xu et~al.(2022)Xu, Liu, and Song]{xu2022functions}
Xu, B., L.~Liu, and G.~Song, 2022.
\newblock Functions and regulation of translation elongation factors.
\newblock \emph{Frontiers in Molecular Biosciences} 8:816398.

\bibitem[Marintchev(2012)]{marintchev2012fidelity}
Marintchev, A., 2012.
\newblock Fidelity and quality control in gene expression.
\newblock Academic Press.

\bibitem[Kim and Sauro(2011)]{kim2011measuring}
Kim, K.~H., and H.~M. Sauro, 2011.
\newblock Measuring retroactivity from noise in gene regulatory networks.
\newblock \emph{Biophysical journal} 100:1167--1177.

\bibitem[Tierrafr{\'\i}a et~al.(2022)Tierrafr{\'\i}a, Rioualen, Salgado, Lara,
  Gama-Castro, Lally, G{\'o}mez-Romero, Pe{\~n}a-Loredo, L{\'o}pez-Almazo,
  Alarc{\'o}n-Carranza, et~al.]{tierrafria2022regulondb}
Tierrafr{\'\i}a, V.~H., C.~Rioualen, H.~Salgado, P.~Lara, S.~Gama-Castro,
  P.~Lally, L.~G{\'o}mez-Romero, P.~Pe{\~n}a-Loredo, A.~G. L{\'o}pez-Almazo,
  G.~Alarc{\'o}n-Carranza, et~al., 2022.
\newblock RegulonDB 11.0: Comprehensive high-throughput datasets on
  transcriptional regulation in Escherichia coli K-12.
\newblock \emph{Microbial Genomics} 8.

\bibitem[Milo et~al.(2010)Milo, Jorgensen, Moran, Weber, and
  Springer]{milo2010bionumbers}
Milo, R., P.~Jorgensen, U.~Moran, G.~Weber, and M.~Springer, 2010.
\newblock BioNumbers—the database of key numbers in molecular and cell
  biology.
\newblock \emph{Nucleic acids research} 38:D750--D753.

\bibitem[Gong et~al.(2008)Gong, Fan, Bilderbeck, Li, Pang, and
  Tao]{gong2008comparative}
Gong, X., S.~Fan, A.~Bilderbeck, M.~Li, H.~Pang, and S.~Tao, 2008.
\newblock Comparative analysis of essential genes and nonessential genes in
  Escherichia coli K12.
\newblock \emph{Molecular Genetics and Genomics} 279:87--94.

\bibitem[Zhu and Dai(2019)]{zhu2019maintenance}
Zhu, M., and X.~Dai, 2019.
\newblock Maintenance of translational elongation rate underlies the survival
  of Escherichia coli during oxidative stress.
\newblock \emph{Nucleic acids research} 47:7592--7604.

\bibitem[Keseler et~al.(2017)Keseler, Mackie, Santos-Zavaleta, Billington,
  Bonavides-Mart{\'\i}nez, Caspi, Fulcher, Gama-Castro, Kothari, Krummenacker,
  et~al.]{keseler2017ecocyc}
Keseler, I.~M., A.~Mackie, A.~Santos-Zavaleta, R.~Billington,
  C.~Bonavides-Mart{\'\i}nez, R.~Caspi, C.~Fulcher, S.~Gama-Castro, A.~Kothari,
  M.~Krummenacker, et~al., 2017.
\newblock The EcoCyc database: reflecting new knowledge about Escherichia coli
  K-12.
\newblock \emph{Nucleic acids research} 45:D543--D550.

\end{thebibliography}

\end{document}